%% file: main_cvpr.tex
\documentclass[10pt,twocolumn,letterpaper]{article}

\usepackage{cvpr}

\input{preamble}

\definecolor{cvprblue}{rgb}{0.21,0.49,0.74}
\usepackage[pagebackref,breaklinks,colorlinks,citecolor=cvprblue]{hyperref}

\def\paperID{6} 
\def\confName{CVPR}
\def\confYear{2024}

\title{Improved Crop and Weed Detection with Diverse Data Ensemble Learning}

\author{Muhammad Hamza Asad$^{1}$, Saeed Anwar$^{2,3}$, and Abdul Bais$^{1}$\\
$^{1}$ University of Regina, Regina SK, Canada \\
$^{2}$ King Fahd University of Petroleum and Minerals, Dhahran, Saudi Arabia\\
$^{3}$SDAIA-KFUPM Joint Research Center for Artificial Intelligence, Dhahran, Saudi Arabia\\
    \tt\small \{maq541,abdul.bais\}@uregina.ca, \tt\small saeed.anwar@kfupm.edu.sa
}

\begin{document}
\maketitle
\input{sec/0_abstract}    
\input{sec/1_intro}
\input{sec/2_relatedworks}

\input{sec/3_methodology}
\input{sec/5_results}
\input{sec/6_conclusion}

{
    \small
    \bibliographystyle{ieeenat_fullname}
    \bibliography{refs}
}

\end{document}

%% file: preamble.tex
\usepackage[dvipsnames]{xcolor}


%% file: sec/0_abstract.tex
\begin{abstract}
Modern agriculture heavily relies on Site-Specific Farm Management practices, necessitating accurate detection, localization, and quantification of crops and weeds in the field, which can be achieved using deep learning techniques. In this regard, crop and weed-specific binary segmentation models have shown promise. However, uncontrolled field conditions limit their performance from one field to the other. To improve semantic model generalization, existing methods augment and synthesize agricultural data to account for uncontrolled field conditions. However, given highly varied field conditions, these methods have limitations. To overcome the challenges of model deterioration in such conditions, we propose utilizing data specific to other crops and weeds for our specific target problem. To achieve this, we propose a novel ensemble framework. Our approach involves utilizing different crop and weed models trained on diverse datasets and employing a teacher-student configuration. By using homogeneous stacking of base models and a trainable meta-architecture to combine their outputs, we achieve significant improvements for Canola crops and Kochia weeds on unseen test data, surpassing the performance of single semantic segmentation models. We identify the UNET meta-architecture as the most effective in this context. Finally, through ablation studies, we demonstrate and validate the effectiveness of our proposed model. We observe that including base models trained on other target crops and weeds can help generalize the model to capture varied field conditions. Lastly, we propose two novel datasets with varied conditions for comparisons. Our code will be available at github.com.
\end{abstract}

%% file: sec/1_intro.tex
\section{Introduction}
\label{sec:intro}
Throughout human history, farming has been the most crucial component of society for the survival of human beings. Modern farming requires more than traditional techniques and hugely relies on Site Specific Farm Management (SSFM) which requires timely and accurate detection, localization and quantification of crop and weeds in the field~\cite{hashemi2022deep}. SSFM suggests varying the fertilizer and herbicide application rate to the field~\cite{mulla2016historical} by mapping the variability of crops and weeds. With the recent advances in deep learning, accuracy for object detection, localization, and quantification in digital images has tremendously enhanced~\cite{wu2020recent}. Semantic segmentation is widely employed due to its ability to accurately determine the boundaries of different plant categories in an image~\cite{divyanth2023two, wang2022review}. 

\begin{figure}[t!]
\begin{center}
\begin{tabular}{c@{ }c@{ }c}

\includegraphics[width=.15\textwidth]{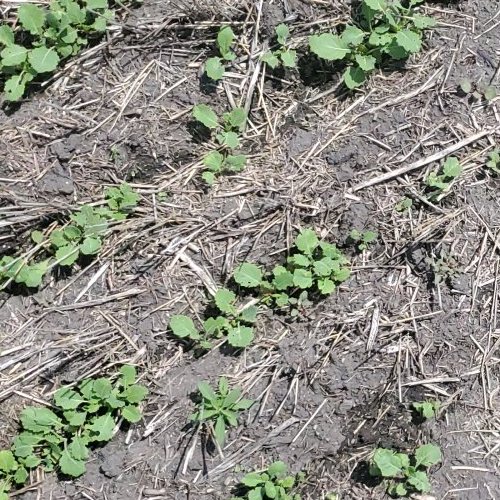}&
\includegraphics[width=.15\textwidth]{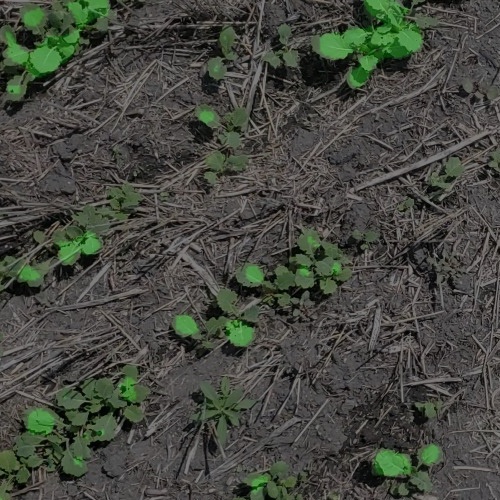}&
\includegraphics[width=.15\textwidth]{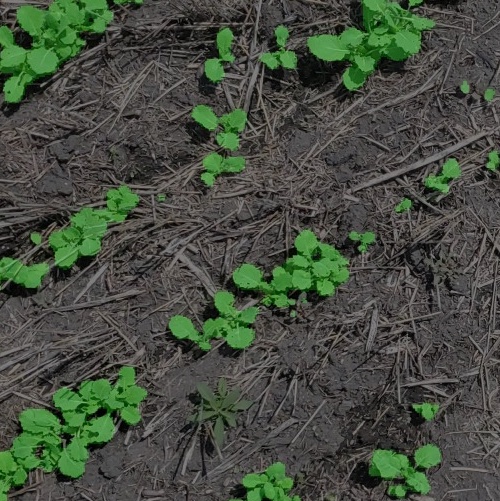}\\

\includegraphics[width=.15\textwidth]{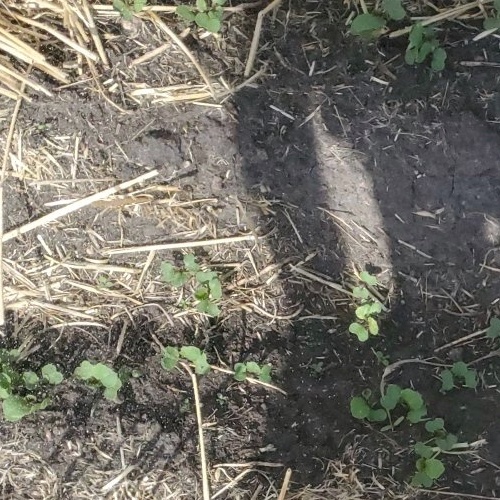}&
\includegraphics[width=.15\textwidth]{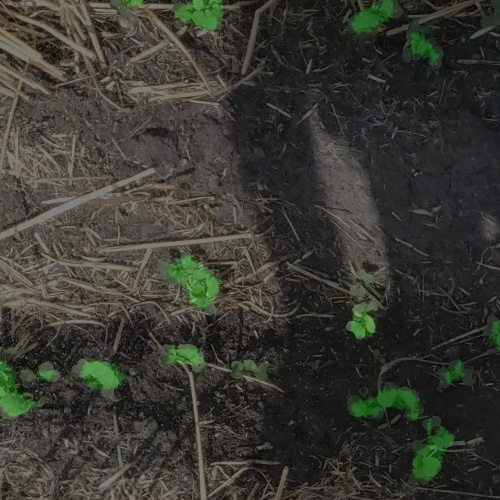}&
\includegraphics[width=.15\textwidth]{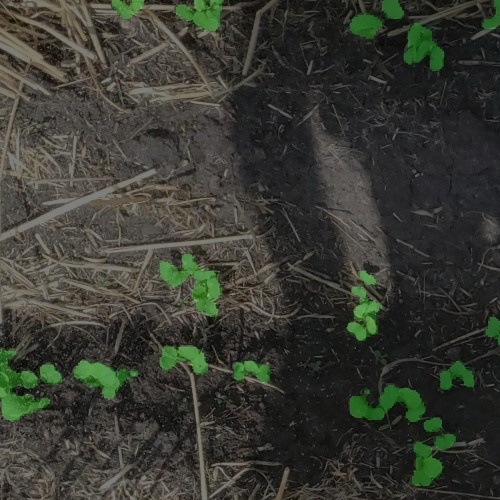}\\
(a) & (b) & (c)\\
\end{tabular}
\end{center}
\caption{(a) Sample images containing early and mid-stage Canola plants, (b) Canola pixels classified by traditional encode-decoder scheme: ResNet50-SegNet. It can be observed that some Canola plant pixels are misclassified as background class. (c) Our proposed framework addresses the false negatives and rightly classifies the majority of Canola pixels. (Best viewed on screen and when zoomed-in).}
\label{fig:placeholder}
\end{figure}

Crop and weed-specific binary semantic segmentation models are commonly employed for improved performance and reducing the effort to manually label every single plant category individually at the pixel level~\cite{asad2019weed, asad2020weed, sapkota2022evaluating}. However, with digital images collected under uncontrolled field conditions, the performance of crop and weed-specific semantic segmentation models deteriorates. Varying image backgrounds, crop staging, crop stress, incidence of unseen vegetation, variable ambient lighting conditions and changing parameters of imaging equipment are common causes of model failure from one field to another. Improving models on test data is an active area of research in semantic segmentation-based models~\cite{chao2021rethinking, hoffman2016fcns, hoffman2018cycada, gong2019dlow, wu2018dcan, zheng2019unsupervised, choi2019self, zheng2021rectifying, tranheden2021dacs}. The challenges presented by agriculture imagery collected under uncontrolled field conditions warrant developing solutions for model generalization. 

Few works in the literature \cite{lu2022generative, shkanaev2020unsupervised, ma2022multi, kwak2022unsupervised} apply domain adaptation to the semantic segmentation models bridging the domain gap by augmenting labelled data using Generative Adversarial Networks (GANs)~\cite{lu2022generative}. Other methods include adversarial domain adaptation, where a model learns features robust to domain changes~\cite{shkanaev2020unsupervised, ma2022multi, kwak2022unsupervised}. In agriculture applications, image enhancement methods are applied to augment data to generalize semantic segmentation models~\cite{zou2021modified, su2021data, wang2020semantic}. However, only some works~\cite{sapkota2022evaluating} achieved generalization by learning from the data of other crops and weeds. 

We develop a novel method that employs an ensemble framework to achieve generalization as models trained for a different target (crop or weed) task can bridge any domain gap. To avoid the need for end-to-end ensemble prediction, we use teacher-student configuration to train the student model from an ensemble of crop and weed models. In such a setting, two ensemble strategies are used: heterogeneous and homogeneous ensemble~\cite{chao2021rethinking}. We opt for homogeneous stacking of the base models utilizing different crop models trained on diversified datasets. As each base model is trained on a different crop, fusing the output of the base teacher models is performed using a trainable meta-architecture. Using this methodology, we improve the mean Intersection Over Union (mIOU) for the Canola crop by up to 12\% and for Kochia weed by 6\% on unseen test data compared to single ResNet50-SegNet semantic segmentation. We also observe that the UNET meta-architecture performs better than other meta-architectures. 

We claim the following three contributions.
\begin{itemize}
    \item We introduce homogeneous stacking of different crop and weed models, which is not investigated before.
    \item We propose a novel knowledge distillation framework which ensembles different crop/weed teacher models using semantic meta-architecture.
    \item We evaluate the proposed model by performing different ablation studies.
\end{itemize}

%% file: sec/2_relatedworks.tex
\section{Related Work}
\label{sec:lit}
Currently, the encoder-decoder framework is widely employed in semantic segmentation for crop and weed detection~\cite{hu2021deep}. UNET is such a network that has symmetrical layers of encoder and decoder. Feature maps from each encoder layer are connected to the corresponding decoder layer~\cite{ronneberger2015u}. Deep learning architectures like VGG~\cite{simonyan2014very} and ResNet50 are used as the encoder block with UNet. Another widely employed encoder-decoder architecture is SegNet~\cite{badrinarayanan2017segnet}. In SegNet, indices of pooling layers are transferred from encoder layers to corresponding decoder upsampling layers. Using these networks, mIOU of 82\% is achieved for weed mapping in Canola fields~\cite{asad2020weed}. UNet architectures are tailored for agriculture data to improve crop and weed mapping~\cite{ullah2021end, zou2021modified}. These encoder-decoder frameworks' performance decreases with changing scale of the objects and reduced feature resolution. To address these challenges, recently, DeepLab~\cite{chen2017deeplab} has been used for crop and weed discrimination~\cite{hashemi2022deep, yu2022development}. Though the object scale changes problem is addressed, these crop and weed detection methods are sensitive to field conditions. With a slight domain shift, the model's performance deteriorates. To bridge domain gaps, augmentation methods are mainly employed in agriculture data~\cite{zou2021modified, su2021data, wang2020semantic}. These methods include random image cropping and patching~\cite{su2021data, zou2021modified}, image enhancement techniques~\cite{wang2020semantic}, and traditional augmenters applied to agriculture images~\cite{ullah2021end}. However, synthetic and augmented data do not account for diverse scenarios of real field conditions. 

In deep learning, ensemble methods are widely employed to improve model generalization. However, the ensemble of semantic segmentation models for crop and weed detection is overlooked. To improve model generalization for uncontrolled field conditions, ensemble methods can be adapted to agriculture data. Ensemble learning reduces overall bias and variance by using the collective knowledge of the base models to make predictions. Ensemble learning can be divided into two distinct steps, namely, ensemble strategy and fusion strategy~\cite{GANAIE2022105151}. Next, we discuss these strategies in detail.

\subsection{Ensemble Strategies}
Ensemble strategies deal with training base models to achieve diversity. Three commonly employed ensemble strategies are 1) bagging, 2) boosting, and 3) stacking. 

\vspace{1mm}
\noindent
\textbf{Bagging methods} generate multiple bags of training data to train base models \cite{breiman2001random}. Bagging in machine learning is employed in techniques like SVM, neural networks and stacked denoising autoencoders~\cite{khwaja2015improved, alvear2018building}. Bagging is helpful in addressing challenging problems such as over-fitting and data imbalance ~\cite{tao2006asymmetric, blaszczynski2015neighbourhood}.

\vspace{1mm}
\noindent
\textbf{Boosting Technique:} In boosting to make predictions, weak learners are trained with equal weights given to each instance. In subsequent training sessions, more weight is given to the misclassified instances so weak learners can learn from challenging cases. AdaBoost and gradient boosting are commonly employed boosting methods~\cite{freund1996experiments, friedman2001greedy}. Recently, boosting has been used with deep neural networks to improve the generalization of the models. Deep belief network, deep boost, incremental boosting CNN, stage-wise boosting CNN, and snapshot boosting are found in the literature to improve the effectiveness and efficiency of boosting methods~\cite{liu2014facial, kuznetsov2014multi, han2016incremental, walach2016learning, zhang2020snapshot}.

\vspace{1mm}
\noindent
\textbf{Stacking strategy}: trains a meta-architecture through distinct architectures, algorithms or hyper-parameter settings and combines the base learners' outputs. Welchowski~\etal~\cite {welchowski2016framework} and Wang~\etal~\cite{wang2020particle} improved generalization and reduced bias through different variants of deep convex nets and deep stacking networks, respectively. In medical imaging, Das~\etal~\cite{das2022deep} operate different encoder-decoder architectures like SegNet~\cite{badrinarayanan2017segnet} and UNet~\cite{ronneberger2015u} to classify brain tumours through semantic segmentation. In some cases, features at different stages of the network are also extracted and fused for improved performance of the deep neural network~\cite{chen2021efcnet}.

Apart from the above categorization of ensemble strategies, another categorization is homogeneous and heterogeneous ensemble learning~\cite{GANAIE2022105151}. As the name indicates, the homogeneous ensemble model uses the same base models; however, to create diversity in the prediction of the base models, randomness and uncertainty are added to the training data of each base model. On the other hand, Bagging is an example of homogeneous ensemble learning. Contrary to it, heterogeneous ensemble learning uses multiple architectures as a base model with varying computational costs~\cite{kilimci2018deep}. The output of these base learners is fused for final prediction. Following subsection details ensemble learning output fusion strategies.

\subsection{Fusion Strategies} 
The final prediction of an ensemble model depends on the approach to fuse the outputs. Multiple strategies can achieve this goal, such as averaging, majority voting, \etc. Although averaging is simple since it's not adaptive, the outcome usually needs improvement~\cite{GANAIE2022105151}. Similarly, majority voting performs well for shallower networks compared to deep neural networks~\cite{ju2018relative}. Further, a trainable meta-layer is also a commonly applied method for finding the weight of base models in the final output of the ensemble. Stacked generalization and super learner methods are also widely used for regression and classification problems~\cite{wolpert1992stacked, ju2019propensity}.

Our method adapts above mentioned ensemble strategies to semantic segmentation applications in agriculture. We develop a homogeneous stacking ensemble with a trainable meta-architecture on the top to fuse the output of base models trained on diversified targets (crops and weeds).

%% file: sec/3_methodology.tex
\subsection{Methodology}
\label{subsec:methodology}

\begin{figure*}[!htbp]
         \centering
         \includegraphics[width= 0.90\textwidth]{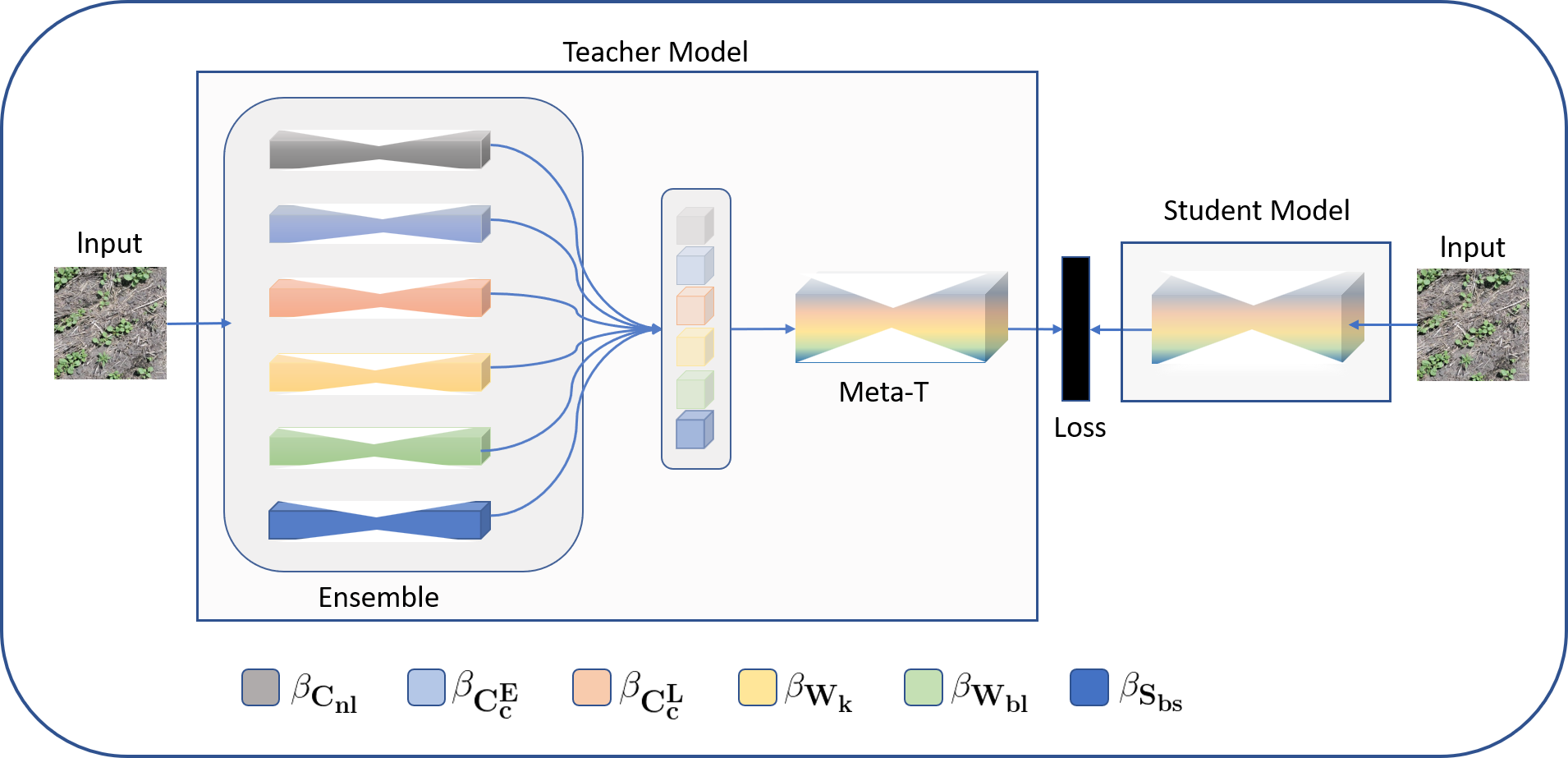}
         \caption{Our proposed framework for the ensemble of the teachers-student model. $\beta$'s are the base models, which are fused using meta-architecture. Table~\ref{CH3:teacher_models} presents the details of base models and respective datasets used for training. The student model is trained for a specific problem (such as Canola) to learn from the ensemble base models trained on different datasets.}
         \label{CH3:meta_arch}
\end{figure*}

Recently, a few attempts have been made to propose semantic segmentation models for crops and their different growth stages, as well as models specific to weeds. Nevertheless, these individual models did not perform well due to changes in field conditions, \eg, ambient lighting. Also, the occurrence of new unseen vegetation types, background soil, and crop residue may fail the model on new fields. Given the availability of data for different target problems, we investigate if the individual train on specific data can be used to account for varying field conditions. Therefore, a homogeneous stacking ensemble of base models trained on different datasets is proposed. Such an ensemble strategy requires a fusion different than simple averaging. Addressing the generalizing semantic segmentation models using ensemble learning may result in a computationally costly end-to-end ensemble, which warrants training of student models\footnote{We train each student model individually for the specific data type.} from an ensemble of base models in the teacher. Figure~\ref{CH3:meta_arch} illustrates the flow diagram of the proposed framework.  
 
\begin{table*}[t!]
\caption{Details of base models architecture and respective datasets: Each model is trained on a separate target crop/weed/soil.} 
\label{CH3:teacher_models}
\centering 
\resizebox{\textwidth}{!}{
\begin{tabular}{l c l l c}
\hline 
Base Models & Pre-trained & Description & Datasets & No. of Images\\ 
\hline
 $\beta_{C_{nl}}$ & $\checkmark$ & Detects narrow leaf crops & NLD: Narrow Leaf Dataset & 250 \\
 $\beta_{C_{c}^{E}}$ & $\checkmark$ & Detects early stage Canola & ESCCD: Early Stage Canola Crop Dataset & 150 \\
$\beta_{C_{c}^{L}}$ & $\checkmark$ & Detects mid / late stage Canola & LSCCD: Late Stage Canola Crop Dataset & 300 \\
 $\beta_{W_{k}}$ & $\checkmark$ & Detects Kochia weed & KWD: Kochia Weed Dataset & 124 \\
 $\beta_{W_{bl}}$& $\checkmark$ & Detects broad leaf weeds & BLWD: Early Stage Canola Crop Dataset & 150 \\
$\beta_{S_{bs}}$& $\checkmark$ & Detects bare soil & TSD: Total Soil dataset & 50 \\
\hline
\end{tabular}}\par
\end{table*}

\subsection{ArgMax Baseline Ensemble}
ArgMax Baseline Ensemble (ABE) is a rule-based baseline model that compares predictions for a pixel with the predictions of the base models ($\beta$). The meta-T model ($\tau$) classifies pixels based on the decision of the base models, which predicts it more confidently. For any pixel $(x,y)$ and $i$ binary models in an ensemble $f_i = \beta_i(x,y)$, class $C_\tau$ of the $(x,y)$ is given by the following equation:

\begin{equation}
C_\tau = \max( f_i ),
\label{CH3:Eq_argmax}
\end{equation}
where $\max( \cdot)$ is the $\tau$ and $i \in {1,2, \cdots, N}$.
\begin{table}[!tbp]
\caption{The architecture of each proposed model with the ensemble (base models), Meta-T and Student.}
\label{tab:Methods}
\centering 
\resizebox{\columnwidth}{!}{
\begin{tabular}{l |c c| c c|c c}
\hline 
   & \multicolumn{2}{c|}{Base} & \multicolumn{2}{c|}{Meta-T} & \multicolumn{2}{c}{Student}\\ \cline{2-7}
Models & Enco. & Deco. & Enco. & Deco&.Enco. & Deco. \\
\hline 

ABE  & R50 & Convs & \multicolumn{2}{c|}{$max(\cdot)$} &R50 & Convs \\
MCE  & R50 & Convs & \multicolumn{2}{c|}{One Conv} &R50 & Convs \\
MSNE  & R50 & Convs &  & &R50 & Convs \\
MUNE   & R50 & Convs & R50 & Convs &R50 & Convs \\ \hline
\end{tabular}
}
\end{table}

\subsection{Mask Convolution Ensemble (MCE)}
Unlike the baseline teacher model, which uses the ArgMax rule, MCE adds a trainable layer to the combined output of the base models in the ensemble. This layer determines the contribution of each base model's prediction to the pixel class. To achieve this, we utilize a $1\times 1$ convolutional layer $h$ with sigmoid activation $\sigma$ to reduce the prediction masks of the base models in the ensemble to the desired number of classes. It is assumed that adding a trainable layer on top of the base models will help learn a more complex relationship between the outputs of base models and ground truth. 

\begin{equation}
C_\tau = \sigma(h(f_i)).
\label{eq:MCS}
\end{equation}

Here, the Meta-T architecture is represented as $\sigma(h(f_i))$. 

\subsection{Meta-SegNet Ensemble (MSNE)}
The third ensemble method stems from using multi-stage CNN for classification and localization. The mask convolution method adds a $1 \times 1$ convolution layer over the top of base SegNet models, while here, a whole SegNet model $\psi$ is added in the second stage to combine the outputs of base SegNet models in the ensemble. Furthermore, adding a deep network over the top maps the non-linear complex spatial relationships between different objects in the image. While training Meta-SegNet $\psi$ (Meta-T), base model weights are not updated.

\begin{equation}
C_\tau = \psi(f_i)
\label{eq:MSNE}
\end{equation}

\subsection{Meta-UNet Ensemble (MUNE)}
Most works in the literature involving ensemble learning combine the output of different architectures trained on the same data~\citep{GANAIE2022105151}. However, the ensemble learning methods used in this study achieve diversification by combining models trained on different datasets. To benefit from the dataset and architectural diversification, we apply UNet as a meta-architecture over the top of SegNet base architecture. We have trained SegNet base models for different crops and weeds. Therefore, the base models' weights are not updated during the training of Meta-UNet $\phi$ (Meta-T). We hypothesize that using a different architecture for the Meta-T than the base models brings diversity for improved segmentation accuracy. 

\begin{equation}
C_\tau = \phi(f_i)
\label{eq:MUNE}
\end{equation}

\subsection{Student Model}
The objective is to generalize original base models by sharing the distant learned features of uncontrolled field conditions with each other. After training the ensemble base models in a supervised setting, the student model's Meta-T architecture remains the same as the teacher's Meta-T architecture. The student model is trained unsupervised by minimizing the loss between the teacher and student output. 
\[
\text{Loss} = -\frac{1}{N} \sum_{i=1}^{N} [ C_\tau \log(C_s )+ (1 - C_\tau) \log(1 - C_s) ]
\]
where $C_\tau$ and $C_s$ are predictions from teacher and student models, respectively. The architecture of each proposed model for Ensemble, Meta-T and student are shown in Table~\ref{tab:Methods}.

%% file: sec/5_results.tex
\section{Experiments}

In this section, we provide the details of the experimental setup of our proposed method, followed by our collected datasets. Next, we present detailed comparisons among the state-of-the-art techniques and conclude this section by performing an ablation study, where we analyze the impact of each vital component in the proposed framework.

\subsection{Datasets}
The proposed methods are tested on two datasets: the Kochia Weed Dataset (KWD) and the Multi-Stage Canola Dataset (MSCD). 

\begin{figure*}[t!]
\begin{center}
\begin{tabular}{c@{ }c@{ }c@{ }c}

\includegraphics[width=.24\textwidth]{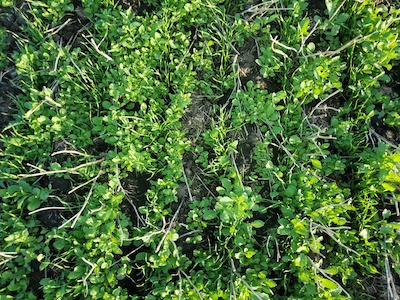}&
\includegraphics[width=.24\textwidth]{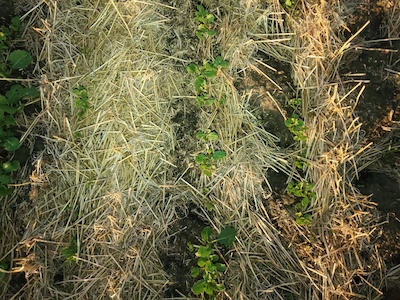}&
\includegraphics[width=.24\textwidth]{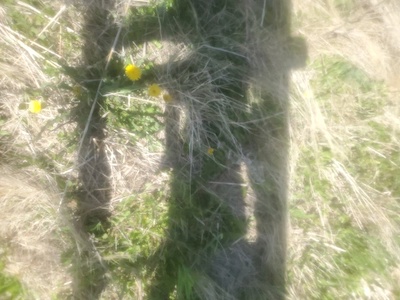}&
\includegraphics[width=.24\textwidth]{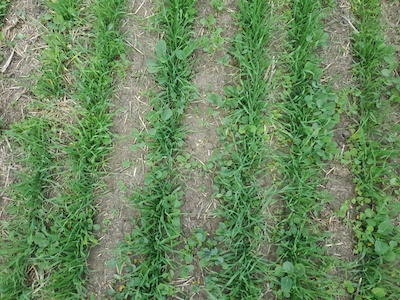}\\
\includegraphics[width=.24\textwidth]{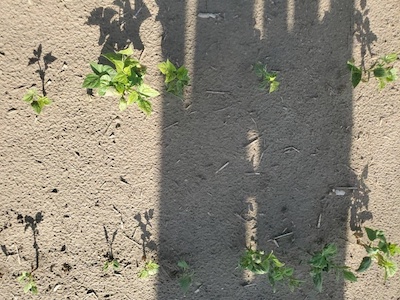}&
\includegraphics[width=.24\textwidth]{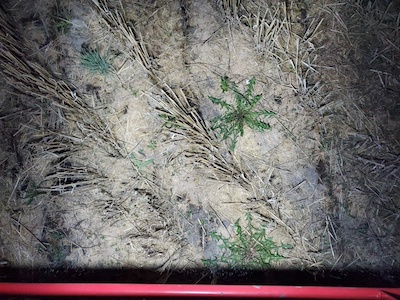}&
\includegraphics[width=.24\textwidth]{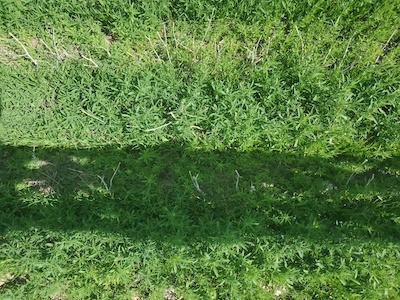}&
\includegraphics[width=.24\textwidth]{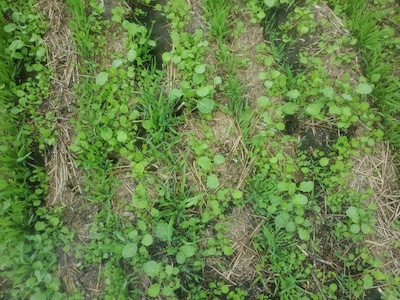}\\
\end{tabular}
\end{center}
\caption{The real world challenging field conditions: the data insight and the challenges posed by uncontrolled field conditions like blurring, variable orientations of crop rows, changing ambient lighting conditions, equipment shadows and images collected during night time under auxiliary lights.}
\label{data_insight}
\end{figure*}

\vspace{1mm}
\noindent
\textbf{Collection Process}: High-resolution RGB images are collected from multiple fields using grid sampling for both datasets. Imaging sensors are mounted on farm machinery, collecting ground images in uncontrolled field conditions. Typically, every image contains 3-5 rows of crop. However, sometimes crop rows swell to 10 due to the increased height of imaging equipment while farm machinery turns at the edges of the field. Notably, the data is collected under uncontrolled field conditions. It includes sunny and cloudy conditions and images during dawn and dusk. Some of the images contain farm machinery shadows. Also, sometimes images are blurry due to mechanical vibration. Figure~\ref{data_insight} provides insight into the data and variations in the images used for this study.

\vspace{1mm}
\noindent
\textbf{KWD-2023}: Kochia weed infests multiple types of crops, like cereal crops and oilseed crops. Our KWD comes from fields of early stage Canola, late stage Canola, Oats, Wheat and Durum. The dataset consists of 99 images.

\vspace{1mm}
\noindent
\textbf{MSCD-2023}: In comparison, Canola data comes from multiple fields, consisting of two stages: early-stage dicots Canola and mid-stage broad leaves Canola. MSCD-2023 contains 305 images acquired in 2023.

However, it is to be noted that base models in the ensemble of the teacher model are trained on separate and distinct datasets. These comprise cereal crops (Oats, Wheat, Barley and Duram), early and mid-stage Canola, broad-leaf weeds, and Kochia weed.

\subsection{Settings}
\noindent
\textbf{Backbones:} Our proposed framework uses ResNet50~\cite{he2016deep} as the backbone encoder block. Furthermore, SegNet~\cite{badrinarayanan2017segnet} architecture is adopted for pre-trained base models in the ensemble. The selection of SegNet in the ensemble models is based on previous works where it performs marginally better than UNet on agriculture data~\cite{asad2019weed, asad2020weed, ma2019fully}. Moreover, the architecture of the student network is the same as the base models, \ie, SegNet. 

\vspace{1mm}
\noindent
\textbf{Setup:} Categorical cross-entropy is the loss function in our proposed framework. The dataset split is 15\%-15\%-70\%  for testing, validation and training. The proposed framework is trained with GPU RTX 3090 support. Adam is used as an optimizer with a learning rate of 0.001, batch size of 2 and input image dimensions of 1440 $\times$ 1088 $\times$ 3. The training dataset is augmented using standard augmenters to avoid overfitting.

\vspace{1mm}
\noindent
\textbf{Metrics:} Classwise IOU, mean IOU and frequency weight IOU (fwIOU) are the performance metrics. Notably, pre-trained base models in the teacher are trained on different datasets particular to the respective target problem - crop/weed.  We assume the diverse data of base models will capture different scenarios of uncontrolled field conditions. 

\subsection{Comparisons}
In this case study, we train the earlier mentioned four learning models: ABE, MCE, MSNE and MUNE. For both Kochia and Canola-specific ensemble models, we use six base models trained  on different datasets and target problems (crop/weed). Our proposed MCE, MSNE and MUNE models are trained and compared with ABE ensemble and  base model $\beta_{W_{k}}$.

\begin{figure*}[t!]
\begin{center}
\begin{tabular}{c@{ }c@{ }c@{ }c}
\includegraphics[width=.3\textwidth]{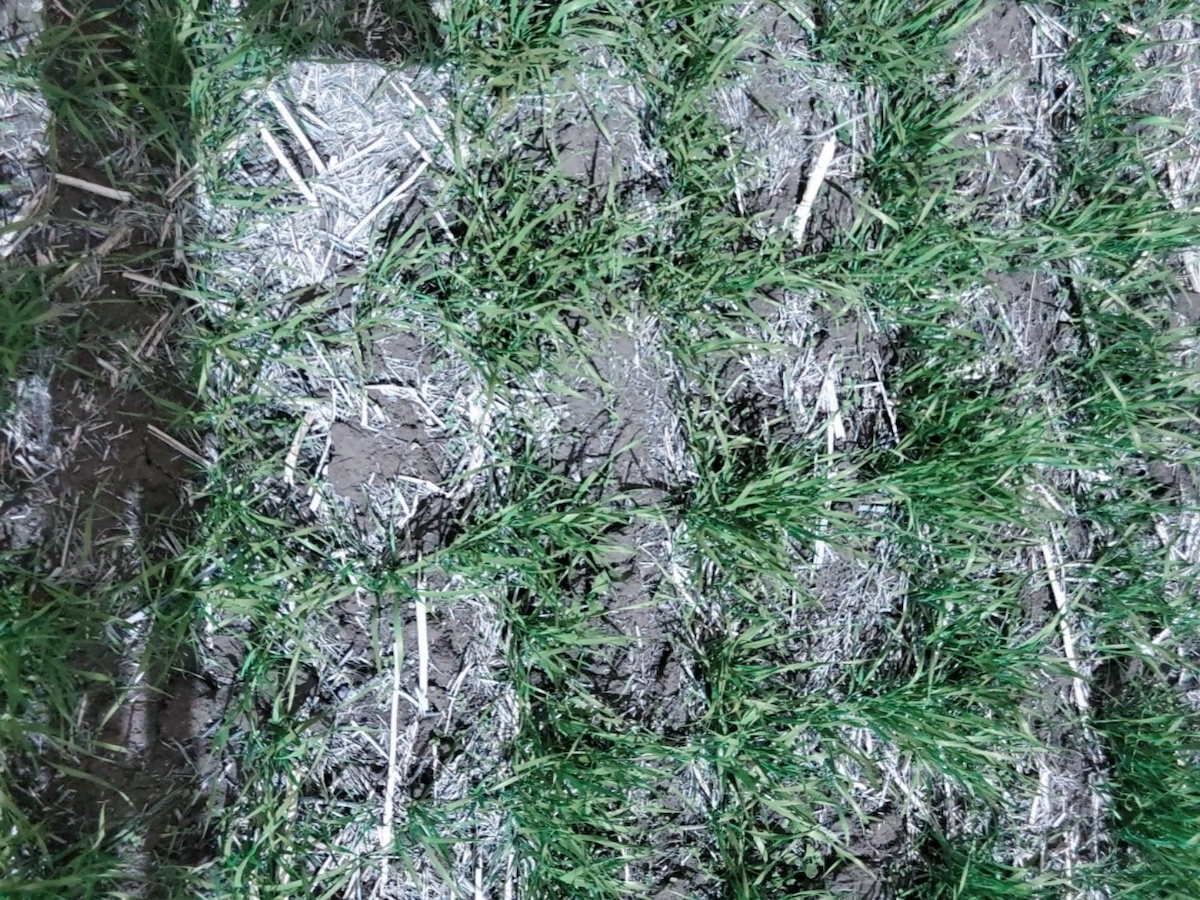}&
\includegraphics[width=.3\textwidth]{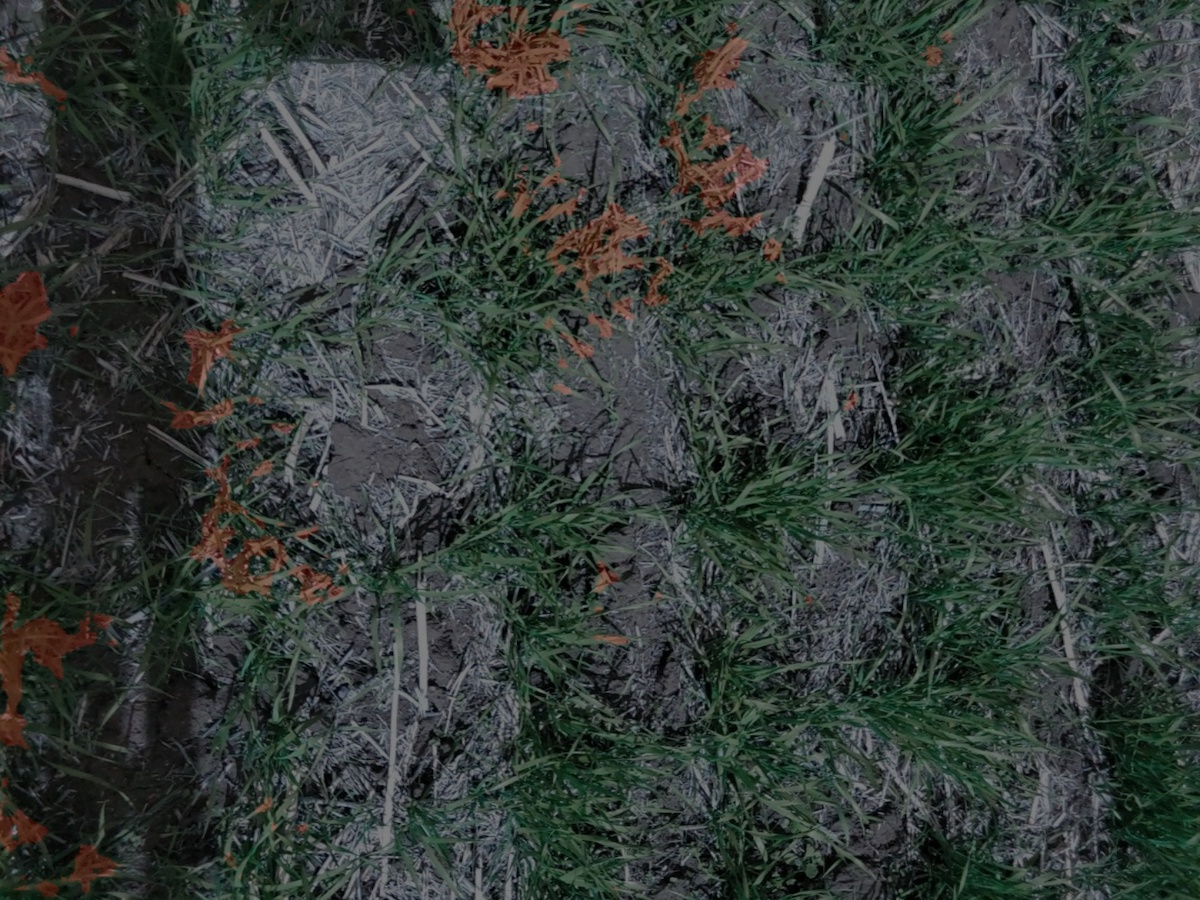}&
\includegraphics[width=.3\textwidth]{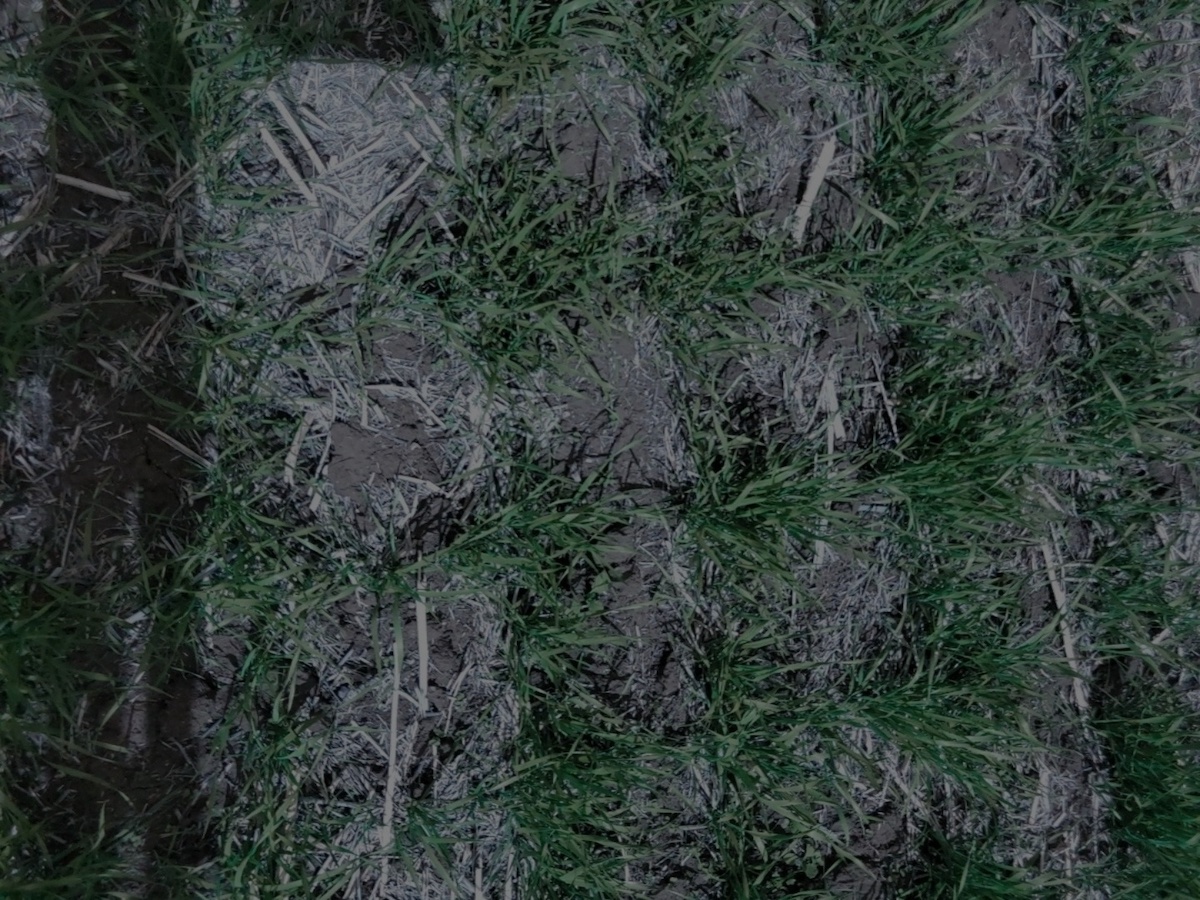}\\

\includegraphics[width=.3\textwidth]{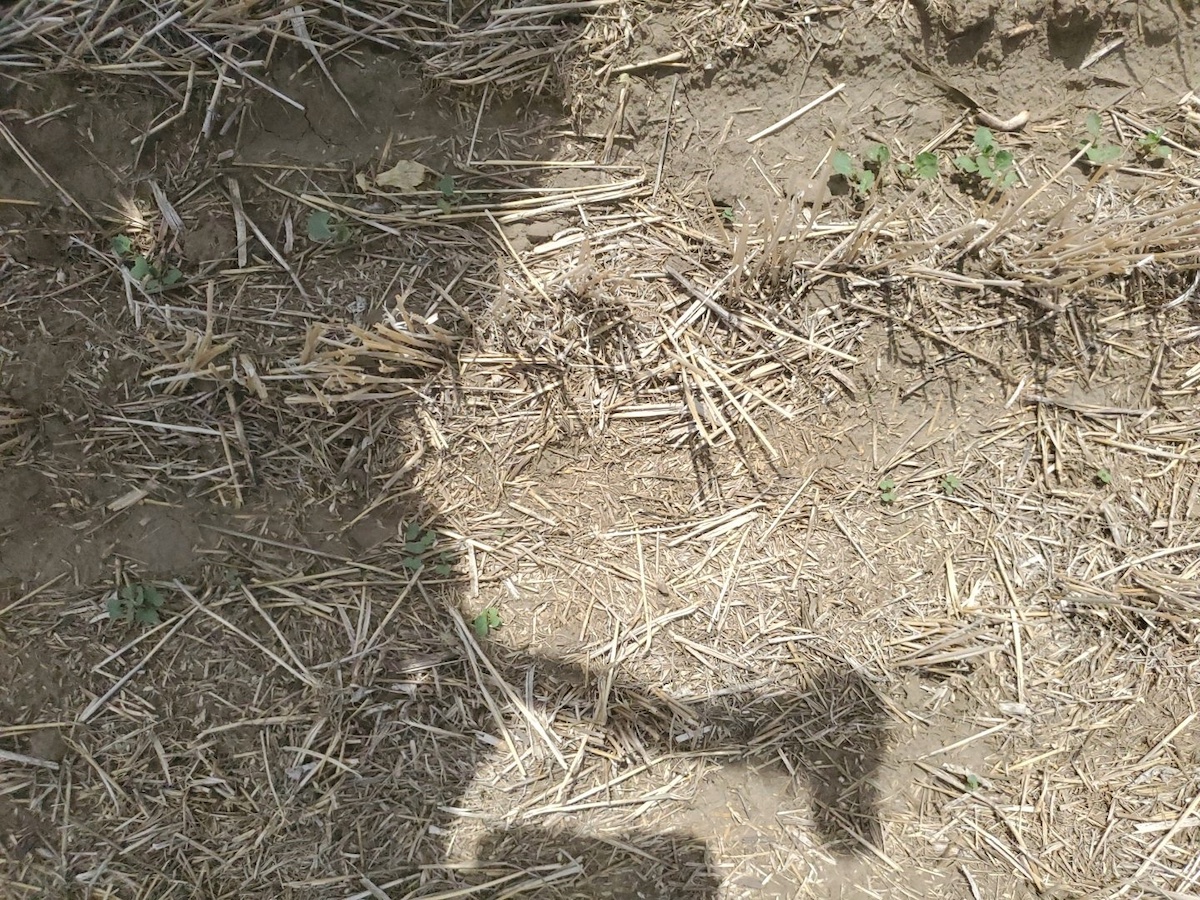}&
\includegraphics[width=.3\textwidth]{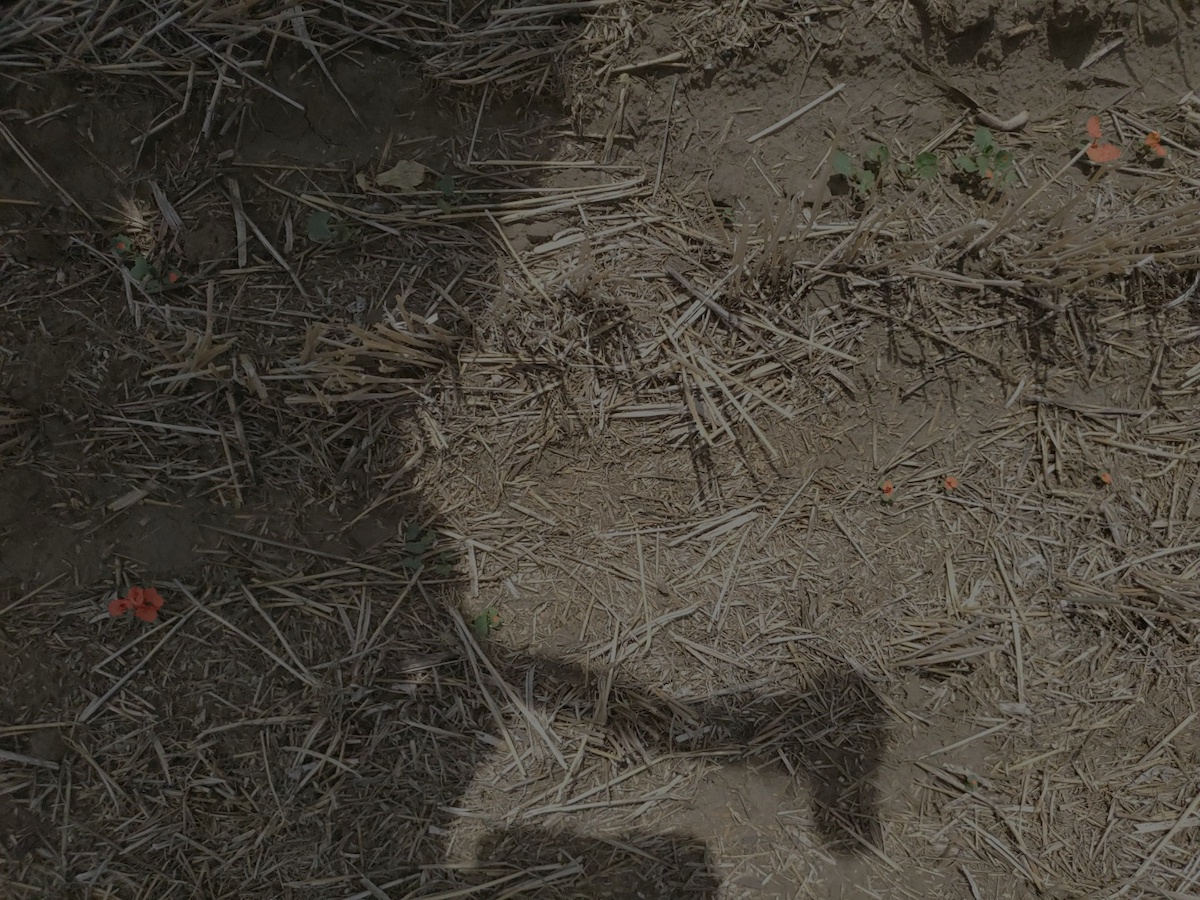}&
\includegraphics[width=.3\textwidth]{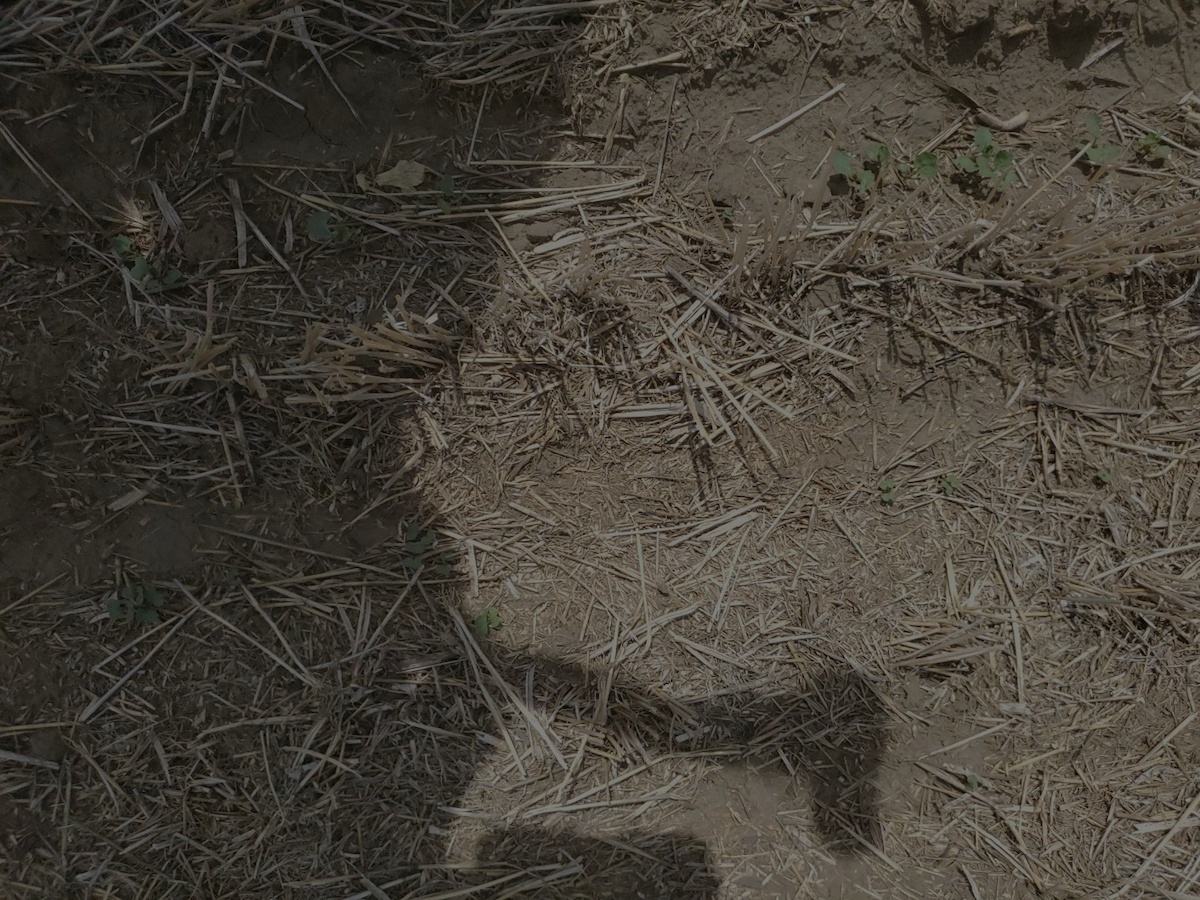}\\
(a) & (b) & (c)\\
\end{tabular}
\end{center}
\vspace{-4mm}
\caption{The visual comparisons from $\beta_{W_{k}}$ and MUNE models. a) The Groundtruth images, b) The prediction of the $\beta_{W_{k}}$ while c)  predictions of MUNE on images. The $\beta_{W_{k}}$ model detects early-stage Canola and narrow leaf as Kochia (false positives), whereas the ensemble model addresses this problem and removes false positives.}
\label{CH3:kochia_ensem_comp}
\end{figure*}

\begin{table}[!tbp]
\caption{Comparing the performance of Kochia-specific deep learning ensemble models. Boldface shows the best-performing model on the test set for the specific metric}
\label{CH3:Kochia_ensem_study3}
\centering 
\resizebox{\columnwidth}{!}{
\begin{tabular}{l c c c c}
\hline  \hline
Models & fwIOU & mIOU & IOU Non-Kochia & IOU Kochia\\
\hline \hline
$\beta_{W_{k}}$ &
0.9256 &
0.8373 &
0.9549 &
0.7198 \\
ABE  &
0.7331 &
0.5769 &
0.7523 &
0.3686  \\
MCE  &
0.9278 &
0.7563 &
0.9573 &
0.5553  \\
MSNE  &
0.9357 &
0.8614 &
0.9604 &
0.7625  \\
MUNE  &
\textbf{0.9371} &
\textbf{0.8638} &
\textbf{0.9615} &
\textbf{0.7638} \\ \hline
\end{tabular}
}\par
\end{table}

\begin{table}[!tbp]
\caption{Comparing the performance of Canola-specific deep learning ensemble models. Boldface shows the best-performing model on the test set for the specific metric.}
\label{CH3:canola_ensem_study3}
\centering 
\resizebox{\columnwidth}{!}{
\begin{tabular}{l c c c c }
\hline \hline
Models & fwIOU & mIOU & IOU Non-Canola & IOU Canola\\
\hline \hline
FCN\_32 & 0.9587 & 0.7136 & 0.9779 & 0.4493\\
PSPNet & 0.9669 & 0.7727 & 0.9821 & 0.5632 \\
UNet & 0.9804 & 0.8628 & 0.9895 & 0.7360\\
SegNet & 0.9766 & 0.8411 & 0.9872 & 0.6950 \\
DeepLab V3+ & 0.9807 & 0.8763 & 0.9858 & 0.7668 \\
HRNet & 0.9832 & 0.8771 & 0.9897 & 0.7643 \\ 
SegFormer & 0.9822 & 0.8835 & 0.9802 & 0.7868\\ \hline
ABE (Ours)  & 0.9406 & 0.8093 & 0.9510 & 0.6676  \\
MCE (Ours)  & 0.9838 & 0.8881 & 0.9913 & 0.7848  \\
MSNE (Ours)  & 0.9811 & 0.8698 & 0.9899 & 0.7497  \\
MUNE (Ours) & \textbf{0.9859} & \textbf{0.9040} & \textbf{0.9923} & \textbf{0.8159} \\ \hline
\end{tabular}}\par
\end{table}

\begin{figure*}[t!]
\begin{center}
\begin{tabular}{c@{ }c@{ }c@{ }c}

\includegraphics[width=.3\textwidth]{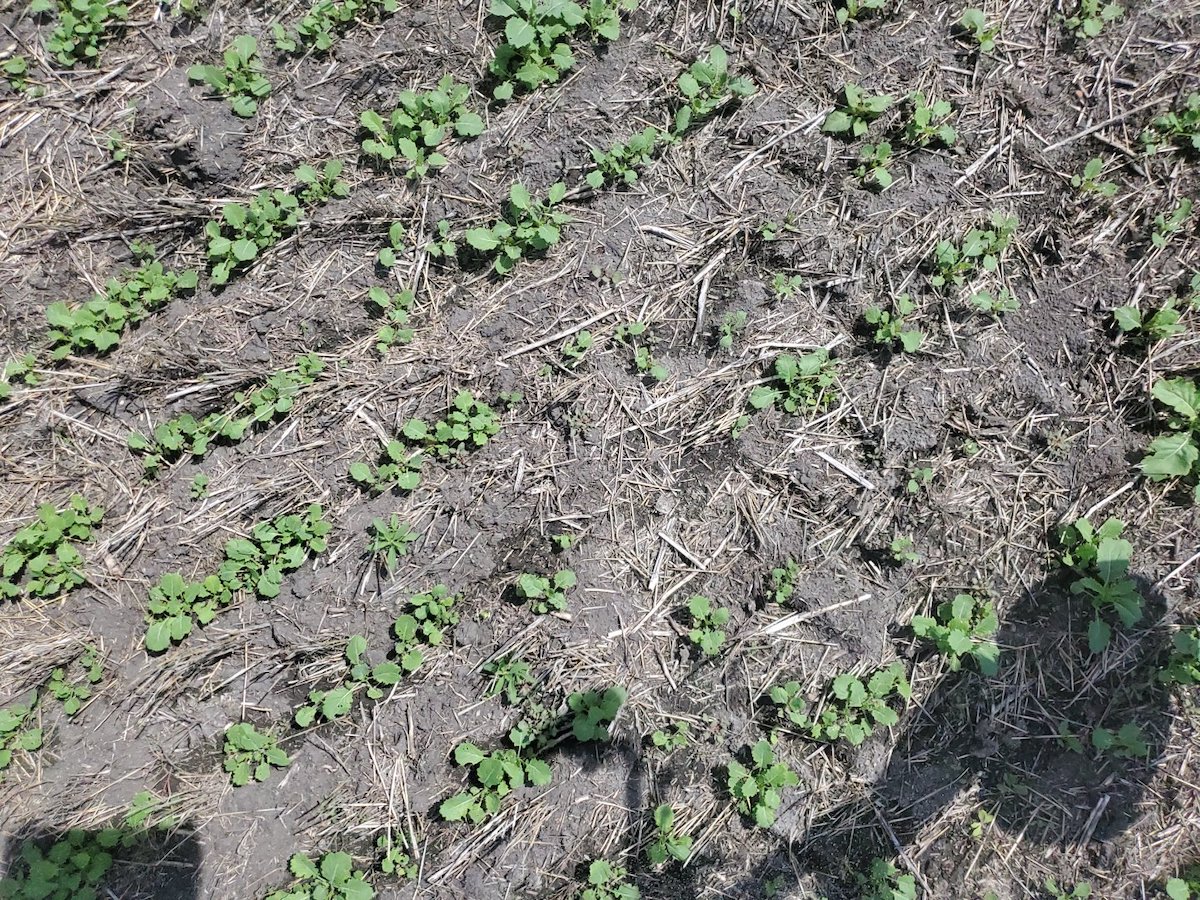}&
\includegraphics[width=.3\textwidth]{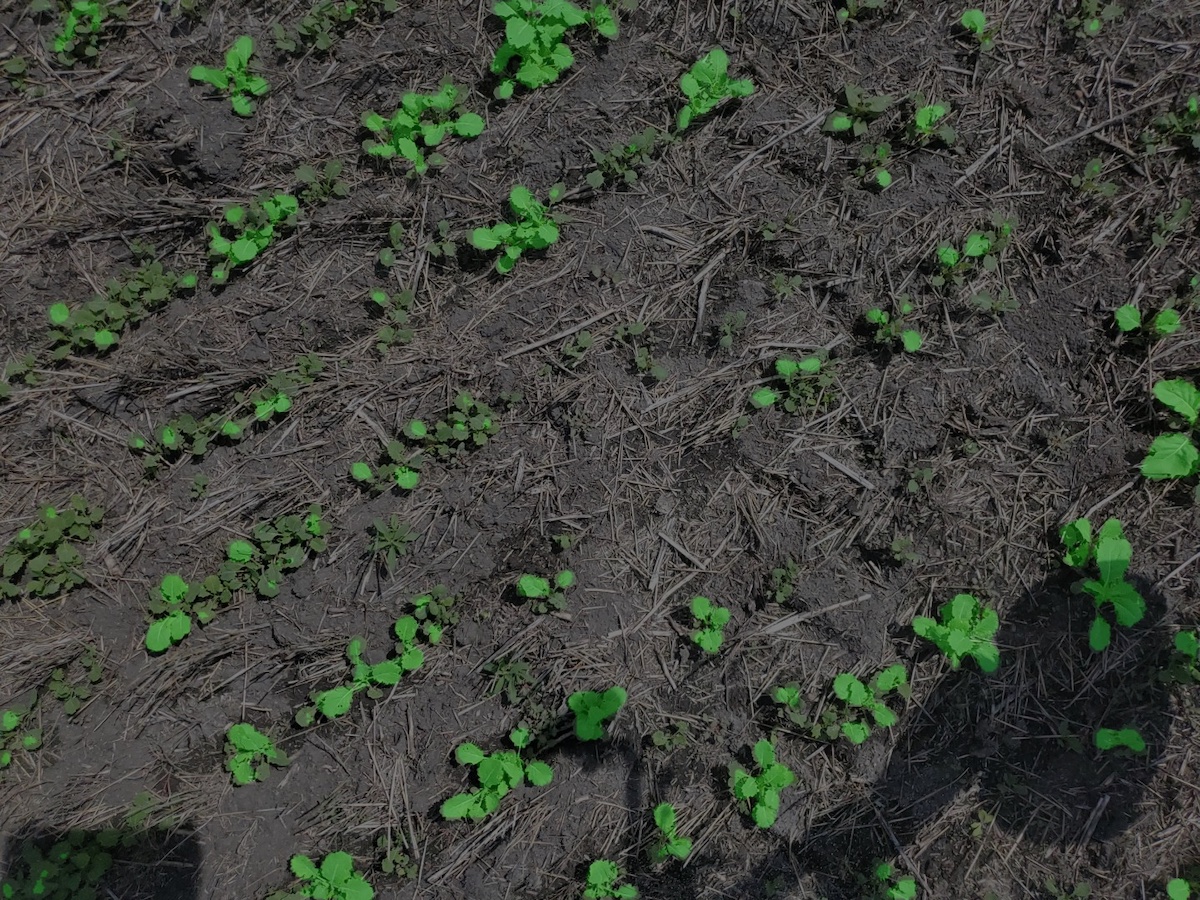}&
\includegraphics[width=.3\textwidth]{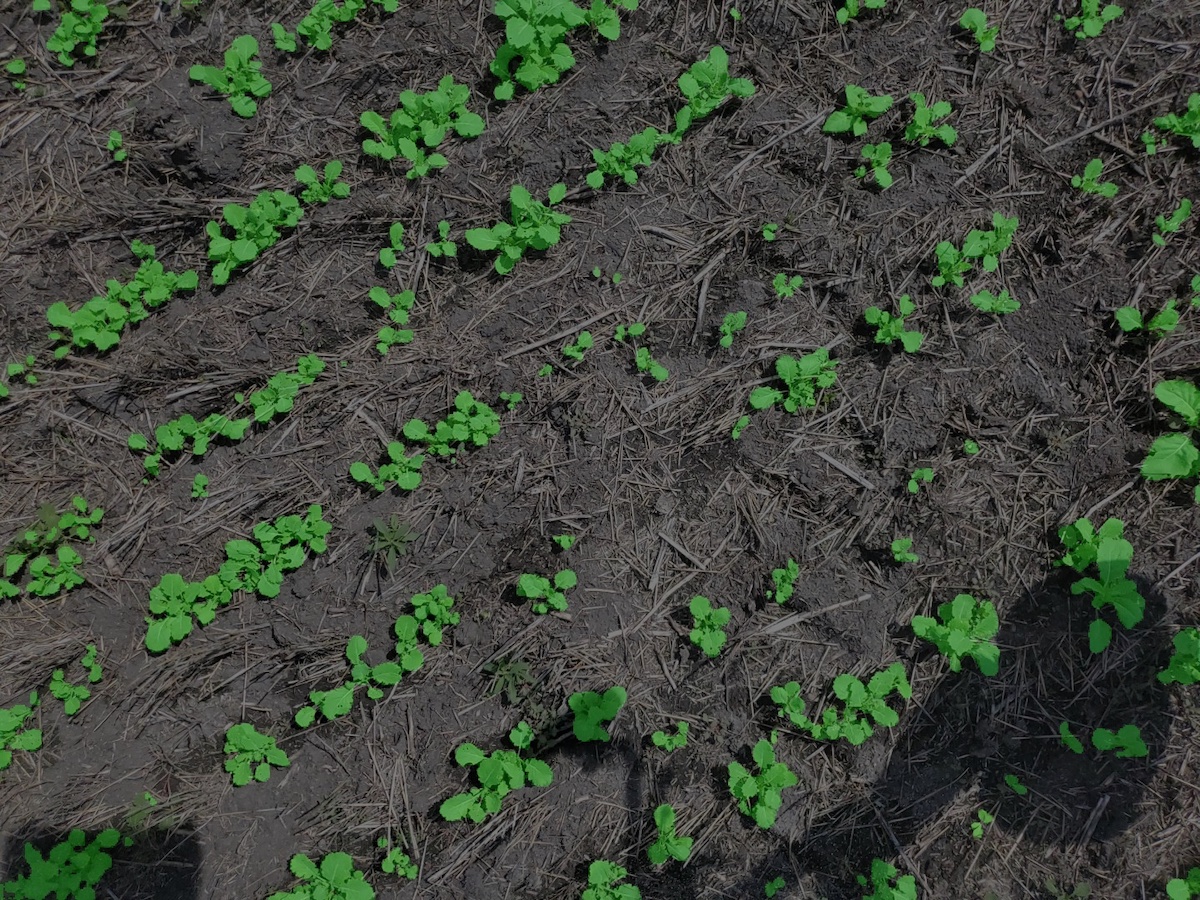}\\

\includegraphics[width=.3\textwidth]{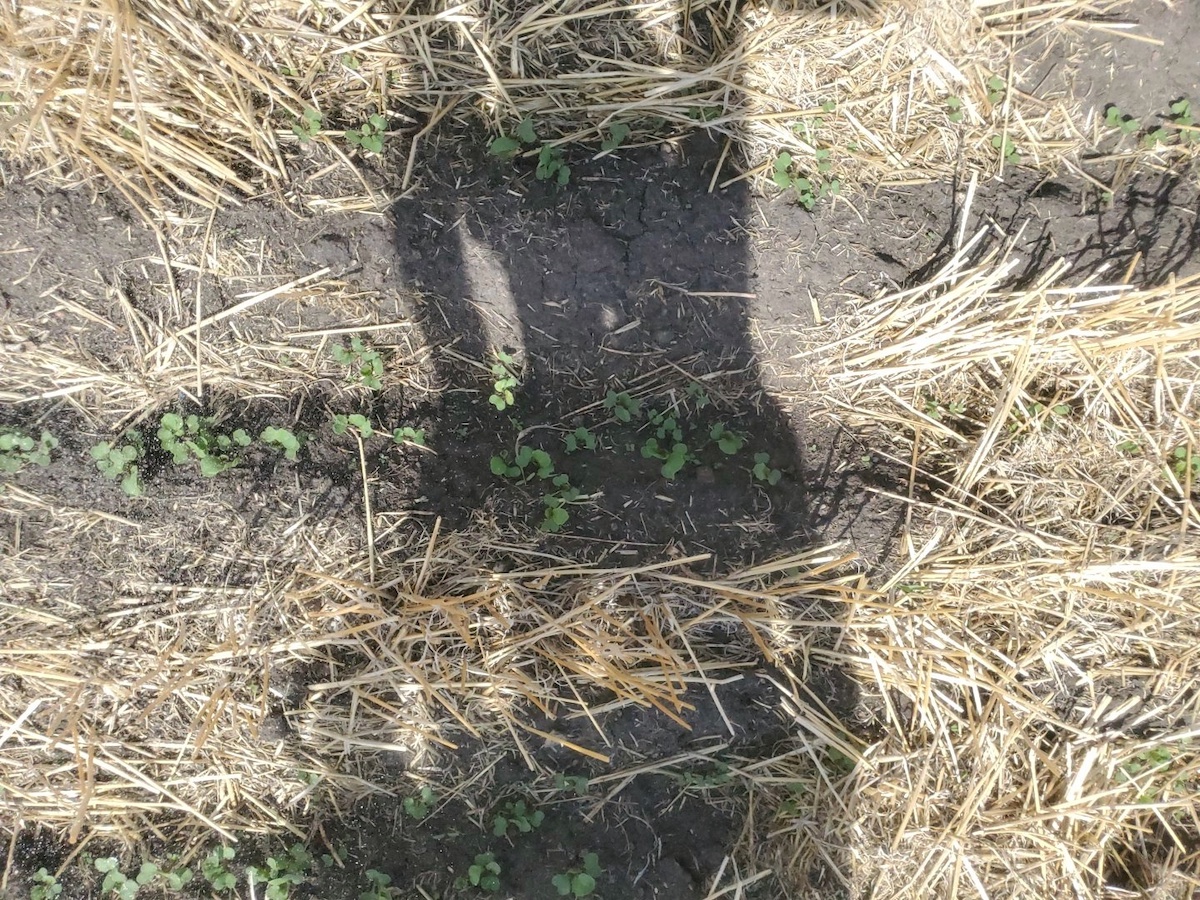}&
\includegraphics[width=.3\textwidth]{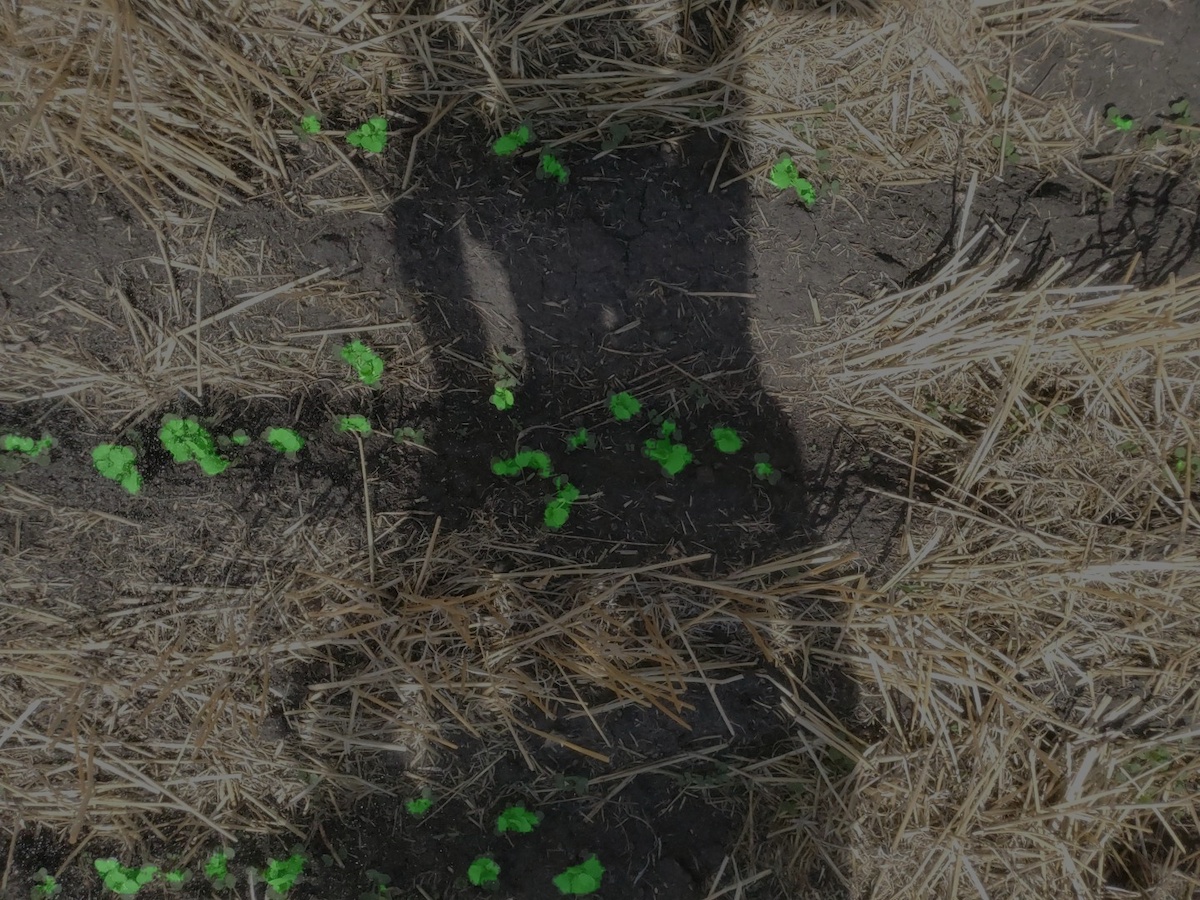}&
\includegraphics[width=.3\textwidth]{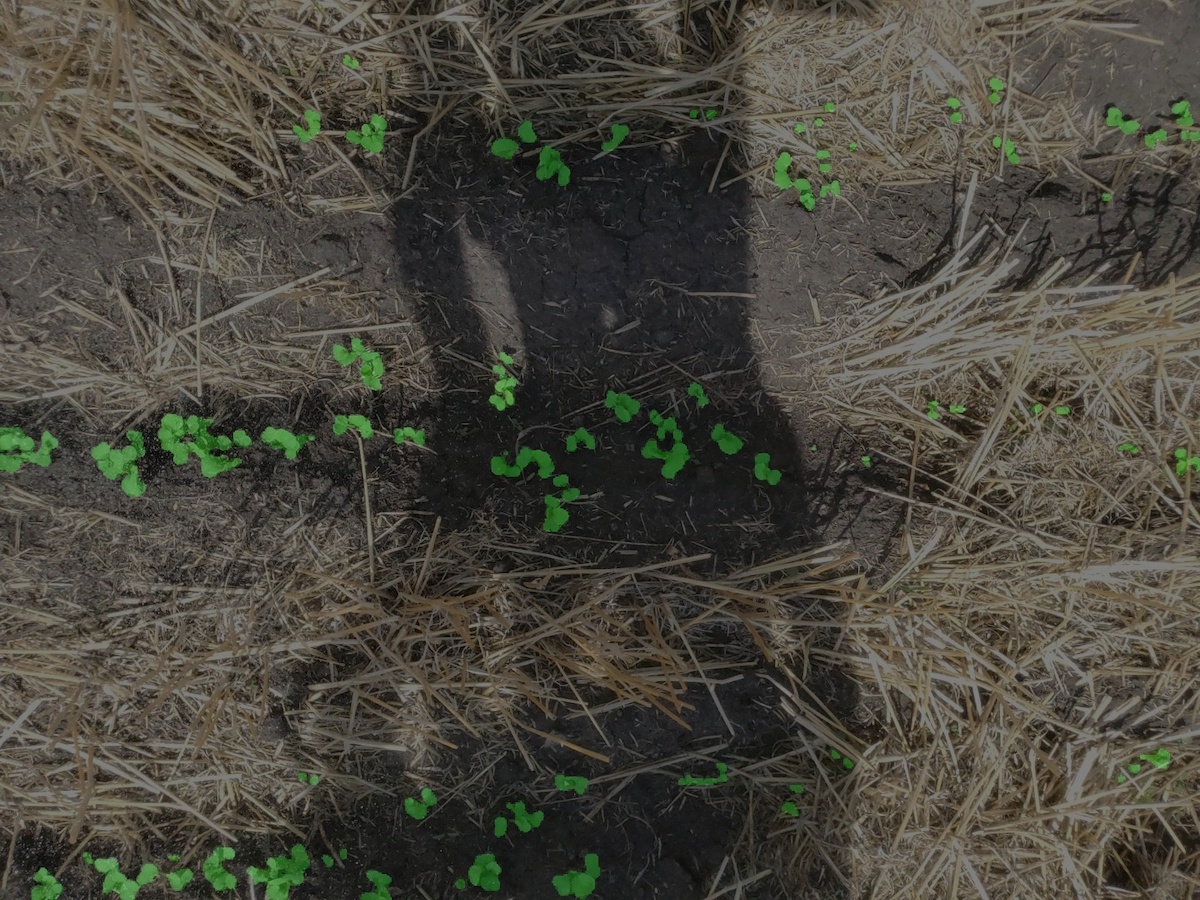}\\
(a) & (b) & (c)\\ 
\end{tabular}
\end{center}
\vspace{-4mm}
\caption{(a) Ground truth, (b) The individual models: $\beta_{C_{c}^{E}}$ \& $\beta_{C_{c}^{L}}$  misses some Canola plants in both early and late stages of the crop (c) our proposed MUNE framework detects missing Canola plants in highly varied field conditions.}

\label{CH3:canola_ensem_seg}
\end{figure*}

\vspace{1mm}
\noindent
\textbf{Comparisons on KWD-2023:} Table~\ref{CH3:Kochia_ensem_study3} presents the results of ensemble methods on KWD-2023. After analyzing the results, we can observe that the MUNE outperforms all other models on all metrics showing slightly higher IOUs than the MSNE. In terms of mIOU, Kochia-specific MUNE performs \emph{2.5\%} better than $\beta_{W_{k}}$. Additionally, Kochia IOU demonstrates a \emph{4.5\%} enhancement compared to $\beta_{W_{k}}$. We also notice that the learnable ensemble meta-architectures perform significantly better than the baseline ABE. Furthermore, multi-class segmentation using one-vs-all binary segmentation models is worse than the original one-vs-all models. Based on the Kochia ensemble models, we can infer that if multi-class labels are available, trainable meta-architectures can combine one-vs-all binary segmentation models for multi-class semantic segmentation with improved IOUs. To illustrate the improvements made by ensemble methods over the $\beta_{W_{k}}$, Figure~\ref{CH3:kochia_ensem_comp} shows some examples and further demonstrate that our proposed model effectively addresses false positive detection of Kochia in Canola and narrow leaf crops via a combination of pre-trained base models. The narrow leaf and Canola models can successfully and confidently detect their respective plants of interest, which helps remove false positive detection of Kochia in the ensemble Kochia settings. Table~\ref{CH3:canola_ensem_study3} demonstrates the ensemble model's effectiveness in detecting all Canola stages by building upon four base models:   $\beta_{C_{nl}}$, $\beta_{C_{c}^{E}}$, $\beta_{C_{c}^{L}}$, and $\beta_{W_{k}}$.

\begin{figure}[t!]
         \centering
         \includegraphics[width= 0.5\textwidth]{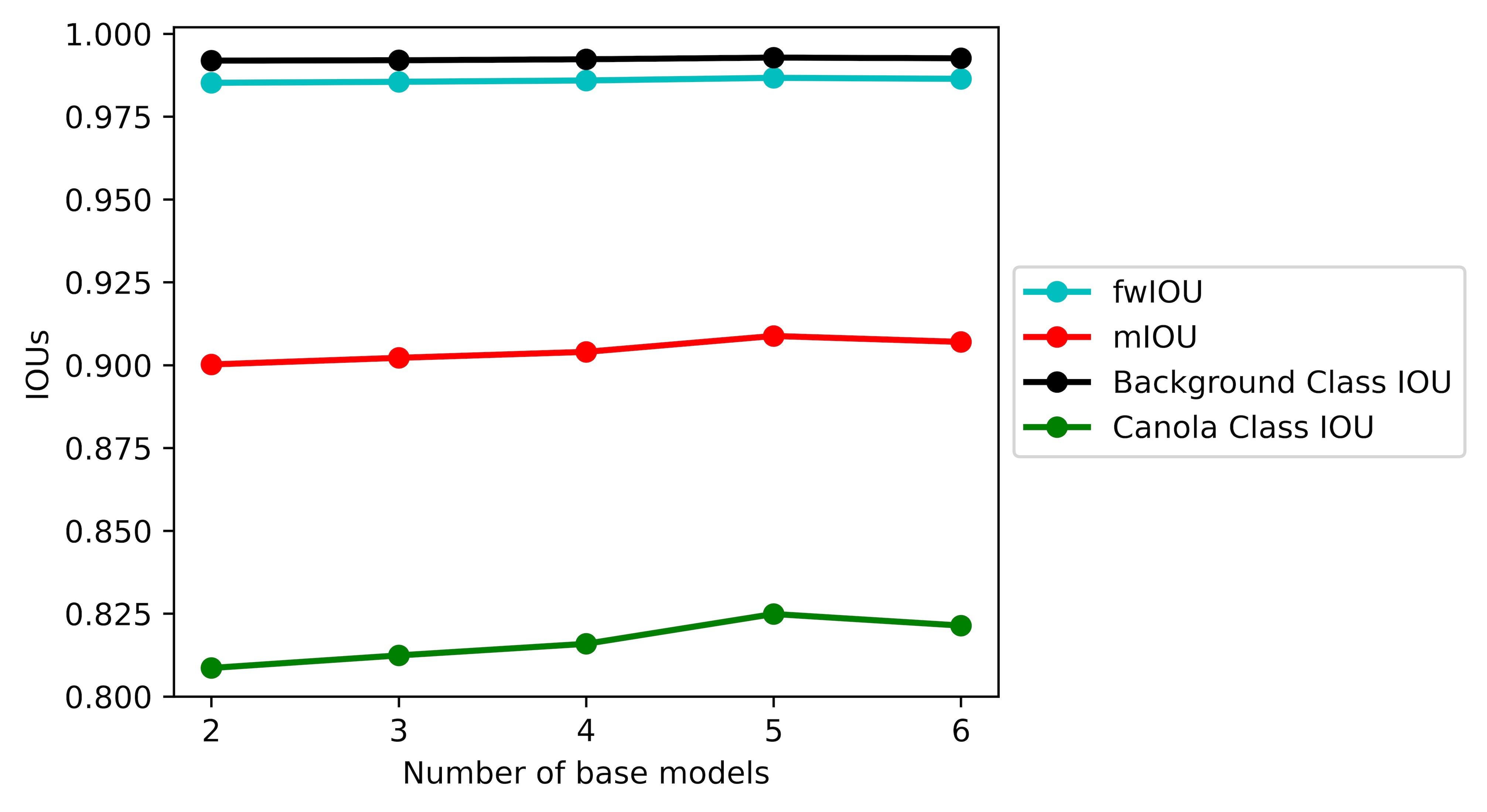}
        \vspace{-2mm}
         \caption{IOUs Vs. the number of base models: It can be observed that increasing the number of base models brings diversification in the ensemble, improving IOUs.}
         \label{ablation_graph}
         \vspace{-2mm}
\end{figure} 

\begin{table*}[t!]
\caption{Comparing the IOU variations with respect to changing the number of base models. The best results are achieved when five base models are used in the ensemble of teachers with the MUNE model.} 
\label{ablation_study1}
\centering 
\begin{tabular}{l|c c c c c c | c c c c }
\hline \hline
Models & $\beta_{C_{c}^{E}}$ &  $\beta_{C_{c}^{L}}$ & $\beta_{W_{k}}$ & $\beta_{C_{nl}}$ & $\beta_{S_{bs}}$ & $\beta_{W_{bl}}$ & fwIOU & mIOU & IOU Non-Canola & IOU Canola\\
\hline 
$M1$ &  \checkmark &  \checkmark  &   &   &   & 
& 0.9852 & 0.9002 & 0.9919 & 0.8086 \\
$M2$ &  \checkmark &  \checkmark  & \checkmark  &   &   & 
& 0.9855 & 0.9022 & 0.9920 & 0.8124\\
$M3$ &  \checkmark &  \checkmark  & \checkmark  & \checkmark  &  &  
& 0.9859 & 0.9040 & 0.9923 & 0.8159\\
$M4$ &  \checkmark &  \checkmark  & \checkmark  & \checkmark  & \checkmark  &  
& \textbf{0.9867}  & \textbf{0.9088} & \textbf{0.9928} & \textbf{0.8249} \\
$M5$ &  \checkmark &  \checkmark  & \checkmark  & \checkmark  & \checkmark  & \checkmark 
& 0.9864 & 0.9070 & 0.9926 & 0.8214 \\
 \hline
\end{tabular}\par
\end{table*}
\vspace{-2mm}
\begin{table*}[tbp]
\caption{Comparing the inference time, Floating Point Operations (FLOPs) and total parameters for end-to-end ensembles.} 
\label{ablation_study2}
\centering 
\begin{tabular}{l|c c c c c c | c c c}
\hline \hline
Models & $\beta_{C_{c}^{E}}$ &  $\beta_{C_{c}^{L}}$ & $\beta_{W_{k}}$ & $\beta_{C_{nl}}$ & $\beta_{S_{bs}}$ & $\beta_{W_{bl}}$ & Inference Time & FLOPs & Parameters \\
\hline 
FCN\_32 & & & & & &  
& 0.80 & 1.28 T & 451 M \\
PSPNet & & & & & &  
& 0.39 & 0.204 T & 29 M \\
UNet & & & & & &  
& 0.50 & 0.572 T & 16 M \\
SegNet & & & & & &  
& 0.48 & 0.428 T & 14 M \\
DeepLab V3+ & & & & & &  
& - & 0.320T & 24 M \\ \hline
$E1$ &  \checkmark &  \checkmark  &   &   &   & 
& 0.50 & 1.86 T & 62 M \\
$E2$ &  \checkmark &  \checkmark  & \checkmark  &   &   & 
& 0.54 & 2.3 T & 77 M \\
$E3$ &  \checkmark &  \checkmark  & \checkmark  & \checkmark  &  &  
& 0.57 & 2.7 T & 92 M \\
$E4$ &  \checkmark &  \checkmark  & \checkmark  & \checkmark  & \checkmark  &  
& 0.61  & 3.1 T & 107 M   \\
$E5$ &  \checkmark &  \checkmark  & \checkmark  & \checkmark  & \checkmark  & \checkmark 
& 0.73 & 3.6 T & 122 M \\
 \hline
\end{tabular}\par
\end{table*}

\vspace{2mm}
\noindent
\textbf{Comparisons on MSCD-2023:} Table~\ref{CH3:canola_ensem_study3} summarises the results of our proposed framework on MSCD-2023. Like KWD 2023, the MUNE model shows the best performance among different ensemble methods with mIOU improvement of 9\% from the ABE model and 5\% from  $\beta_{C_{c}^{E}}$ \& $\beta_{C_{c}^{L}}$ combined. If we compare class-wise Canola IOU, it improves by 14.8\% as compared to the ABE and 12\% from combined Canola models. Figure~\ref{CH3:canola_ensem_seg} presents examples highlighting performance improvements made by our proposed framework. The prediction masks in Figure~\ref{CH3:canola_ensem_seg}(b) are predicted through $\beta_{C_{c}^{E}}$ \& $\beta_{C_{c}^{L}}$ model, while prediction masks on the right are predicted through the Meta-UNet ensemble models. It can be observed that under predictions are made by the original model, as some Canola plants are altogether missed. At the same time, ensemble MUNE significantly addresses the mentioned false negative problem and detects Canola plants to the edges of the crop leaves.

\subsection{Ablation Studies}
In this paper, we perform two types of ablation studies. The first study compares the IOU metrics of MUNE model by varying the number of base models. The second study presents changes in efficiency for end-to-end ensembles with the change in the number of base models. We restrict these ablation studies to MSCD. Figure~\ref{ablation_graph} shows the trend of IOU metrics with respect to the increasing number of base models. We take only those combinations of base models whose target crop/weed is abundant in the field. For example, we included $\beta_{C_{nl}}$ as the base model because it detects narrow leaf crops as well as narrow leaf weeds in the images. It can be observed that IOUs improve with the inclusion of more base models. The peak of IOUs comes when five base models are included in the ensemble of the teacher model. The other reason could be the inclusion of $\beta_{S_{bs}}$ (bare soil extractor) in the ensemble, which helps the model capture different soil backgrounds in the images under variable lighting conditions. However, with the inclusion of the broad leaf base model ($\beta_{W_{bl}}$), the performance drops. It may be due to the confusion caused by $\beta_{W_{bl}}$ with $\beta_{C_{c}^{L}}$ as both models learn features of broad leaves. Notably, a plant type could be a crop in one field, but its occurrence in other fields could be deemed as a weed. Table~\ref{ablation_study1} summarizes the results of the first ablation study.  Similarly, Table~\ref{ablation_study2} presents the inference time, FLOPs and parameters of the end-to-end ensembles. It can be observed that inference time for E1 to E5 models changes from 0.50 seconds to 0.73 seconds, and FLOPs changes from 1.86 T to 3.6 T. Deteriorating efficiency of end-to-end ensemble warrants training student models to improve inference time and require less computational resources.

%% file: sec/6_conclusion.tex
\section{Conclusion}
\label{conc:conclusion}In this paper, we present an ensemble of base models for teacher framework to improve the performance of semantic segmentation models for crop and weed under uncontrolled field conditions. Existing methods attempt to achieve model generalization through augmentation and agriculture data synthesis. However, these methods struggle to capture numerous scenarios of uncontrolled field conditions. To address these challenges, we propose a teacher model trained on diversified target crops and weeds to teach a student model for our target crop/weed. In addition, in the teacher model, we propose a meta-architecture to fuse the outputs of base models trained on different target problems to enhance semantic segmentation performance for crop and weed detection. Our framework will pave the way for research in the cross-applicability of different crop and weed-specific models to \section*{Acknowledgement}
This research project is supported by an NSERC Alliance - Mitacs Accelerate grant (NSERC Ref: ALLRP 581076 - 22 \& Mitacs Ref: IT34971), titled "Developing Machine Learning Methods for RGB Images to Quantify Crop and Weed Populations Across Agricultural Fields" in partnership with Croptimistic Technology Inc.

%% file: main_cvpr.bbl
\begin{thebibliography}{54}
\providecommand{\natexlab}[1]{#1}
\providecommand{\url}[1]{\texttt{#1}}
\expandafter\ifx\csname urlstyle\endcsname\relax
  \providecommand{\doi}[1]{doi: #1}\else
  \providecommand{\doi}{doi: \begingroup \urlstyle{rm}\Url}\fi

\bibitem[Alvear-Sandoval and Figueiras-Vidal(2018)]{alvear2018building}
Ricardo~F Alvear-Sandoval and An{\'\i}bal~R Figueiras-Vidal.
\newblock On building ensembles of stacked denoising auto-encoding classifiers and their further improvement.
\newblock \emph{Information Fusion}, 39:\penalty0 41--52, 2018.

\bibitem[Asad and Bais(2020{\natexlab{a}})]{asad2019weed}
Muhammad~Hamza Asad and Abdul Bais.
\newblock Weed density estimation using semantic segmentation.
\newblock In \emph{Image and Video Technology: PSIVT 2019 International Workshops, Sydney, NSW, Australia, November 18--22, 2019, Revised Selected Papers 9}, pages 162--171. Springer, 2020{\natexlab{a}}.

\bibitem[Asad and Bais(2020{\natexlab{b}})]{asad2020weed}
Muhammad~Hamza Asad and Abdul Bais.
\newblock Weed detection in canola fields using maximum likelihood classification and deep convolutional neural network.
\newblock \emph{Information Processing in Agriculture}, 7\penalty0 (4):\penalty0 535--545, 2020{\natexlab{b}}.

\bibitem[Badrinarayanan et~al.(2017)Badrinarayanan, Kendall, and Cipolla]{badrinarayanan2017segnet}
Vijay Badrinarayanan, Alex Kendall, and Roberto Cipolla.
\newblock Segnet: A deep convolutional encoder-decoder architecture for image segmentation.
\newblock \emph{IEEE transactions on pattern analysis and machine intelligence}, 39\penalty0 (12):\penalty0 2481--2495, 2017.

\bibitem[B{\l}aszczy{\'n}ski and Stefanowski(2015)]{blaszczynski2015neighbourhood}
Jerzy B{\l}aszczy{\'n}ski and Jerzy Stefanowski.
\newblock Neighbourhood sampling in bagging for imbalanced data.
\newblock \emph{Neurocomputing}, 150:\penalty0 529--542, 2015.

\bibitem[Breiman(2001)]{breiman2001random}
Leo Breiman.
\newblock Random forests.
\newblock \emph{Machine Learning}, 45:\penalty0 5--32, 2001.

\bibitem[Chao et~al.(2021)Chao, Cheng, and Lee]{chao2021rethinking}
Chen-Hao Chao, Bo-Wun Cheng, and Chun-Yi Lee.
\newblock Rethinking ensemble-distillation for semantic segmentation based unsupervised domain adaption.
\newblock In \emph{Proceedings of the IEEE/CVF Conference on Computer Vision and Pattern Recognition}, pages 2610--2620, 2021.

\bibitem[Chen et~al.(2021)Chen, Dou, Peng, Li, Sun, and Li]{chen2021efcnet}
Li Chen, Xin Dou, Jian Peng, Wenbo Li, Bingyu Sun, and Haifeng Li.
\newblock {EFCNet}: Ensemble full convolutional network for semantic segmentation of high-resolution remote sensing images.
\newblock \emph{IEEE Geoscience and Remote Sensing Letters}, 19:\penalty0 1--5, 2021.

\bibitem[Chen et~al.(2017)Chen, Papandreou, Kokkinos, Murphy, and Yuille]{chen2017deeplab}
Liang-Chieh Chen, George Papandreou, Iasonas Kokkinos, Kevin Murphy, and Alan~L Yuille.
\newblock Deeplab: Semantic image segmentation with deep convolutional nets, atrous convolution, and fully connected crfs.
\newblock \emph{IEEE PAMI}, 40\penalty0 (4):\penalty0 834--848, 2017.

\bibitem[Choi et~al.(2019)Choi, Kim, and Kim]{choi2019self}
Jaehoon Choi, Taekyung Kim, and Changick Kim.
\newblock Self-ensembling with gan-based data augmentation for domain adaptation in semantic segmentation.
\newblock In \emph{Proceedings of the IEEE/CVF International Conference on Computer Vision}, pages 6830--6840, 2019.

\bibitem[Das et~al.(2022)Das, Bose, Nayak, and Saxena]{das2022deep}
Suchismita Das, Srijib Bose, Gopal~Krishna Nayak, and Sanjay Saxena.
\newblock Deep learning-based ensemble model for brain tumor segmentation using multi-parametric mr scans.
\newblock \emph{Open Computer Science}, 12\penalty0 (1):\penalty0 211--226, 2022.

\bibitem[Divyanth et~al.(2023)Divyanth, Ahmad, and Saraswat]{divyanth2023two}
LG Divyanth, Aanis Ahmad, and Dharmendra Saraswat.
\newblock A two-stage deep-learning based segmentation model for crop disease quantification based on corn field imagery.
\newblock \emph{Smart Agricultural Technology}, 3:\penalty0 100108, 2023.

\bibitem[Freund et~al.(1996)Freund, Schapire, et~al.]{freund1996experiments}
Yoav Freund, Robert~E Schapire, et~al.
\newblock Experiments with a new boosting algorithm.
\newblock In \emph{icml}, pages 148--156. Citeseer, 1996.

\bibitem[Friedman(2001)]{friedman2001greedy}
Jerome~H Friedman.
\newblock Greedy function approximation: a gradient boosting machine.
\newblock \emph{Annals of Statistics}, pages 1189--1232, 2001.

\bibitem[Ganaie et~al.(2022)Ganaie, Hu, Malik, Tanveer, and Suganthan]{GANAIE2022105151}
M.A. Ganaie, Minghui Hu, A.K. Malik, M. Tanveer, and P.N. Suganthan.
\newblock Ensemble deep learning: A review.
\newblock \emph{Engineering Applications of Artificial Intelligence}, 115:\penalty0 105151, 2022.

\bibitem[Gong et~al.(2019)Gong, Li, Chen, and Gool]{gong2019dlow}
Rui Gong, Wen Li, Yuhua Chen, and Luc~Van Gool.
\newblock Dlow: Domain flow for adaptation and generalization.
\newblock In \emph{Proceedings of the IEEE/CVF conference on computer vision and pattern recognition}, pages 2477--2486, 2019.

\bibitem[Han et~al.(2016)Han, Meng, Khan, and Tong]{han2016incremental}
Shizhong Han, Zibo Meng, Ahmed-Shehab Khan, and Yan Tong.
\newblock Incremental boosting convolutional neural network for facial action unit recognition.
\newblock \emph{Advances in Neural Information Processing Systems}, 29, 2016.

\bibitem[Hashemi-Beni et~al.(2022)Hashemi-Beni, Gebrehiwot, Karimoddini, Shahbazi, and Dorbu]{hashemi2022deep}
Leila Hashemi-Beni, Asmamaw Gebrehiwot, Ali Karimoddini, Abolghasem Shahbazi, and Freda Dorbu.
\newblock Deep convolutional neural networks for weeds and crops discrimination from uas imagery.
\newblock \emph{Frontiers in Remote Sensing}, 3:\penalty0 755939, 2022.

\bibitem[He et~al.(2016)He, Zhang, Ren, and Sun]{he2016deep}
Kaiming He, Xiangyu Zhang, Shaoqing Ren, and Jian Sun.
\newblock Deep residual learning for image recognition.
\newblock In \emph{Proceedings of the IEEE conference on computer vision and pattern recognition}, pages 770--778, 2016.

\bibitem[Hoffman et~al.(2016)Hoffman, Wang, Yu, and Darrell]{hoffman2016fcns}
Judy Hoffman, Dequan Wang, Fisher Yu, and Trevor Darrell.
\newblock Fcns in the wild: Pixel-level adversarial and constraint-based adaptation.
\newblock \emph{arXiv preprint arXiv:1612.02649}, 2016.

\bibitem[Hoffman et~al.(2018)Hoffman, Tzeng, Park, Zhu, Isola, Saenko, Efros, and Darrell]{hoffman2018cycada}
Judy Hoffman, Eric Tzeng, Taesung Park, Jun-Yan Zhu, Phillip Isola, Kate Saenko, Alexei Efros, and Trevor Darrell.
\newblock Cycada: Cycle-consistent adversarial domain adaptation.
\newblock In \emph{International conference on machine learning}, pages 1989--1998. Pmlr, 2018.

\bibitem[Hu et~al.(2021)Hu, Wang, Coleman, Bender, Yao, Zeng, Song, Schumann, and Walsh]{hu2021deep}
Kun Hu, Zhiyong Wang, Guy Coleman, Asher Bender, Tingting Yao, Shan Zeng, Dezhen Song, Arnold Schumann, and Michael Walsh.
\newblock Deep learning techniques for in-crop weed identification: A review.
\newblock \emph{arXiv preprint arXiv:2103.14872}, 2021.

\bibitem[Ju et~al.(2018)Ju, Bibaut, and van~der Laan]{ju2018relative}
Cheng Ju, Aur{\'e}lien Bibaut, and Mark van~der Laan.
\newblock The relative performance of ensemble methods with deep convolutional neural networks for image classification.
\newblock \emph{Journal of Applied Statistics}, 45\penalty0 (15):\penalty0 2800--2818, 2018.

\bibitem[Ju et~al.(2019)Ju, Combs, Lendle, Franklin, Wyss, Schneeweiss, and van~der Laan]{ju2019propensity}
Cheng Ju, Mary Combs, Samuel~D Lendle, Jessica~M Franklin, Richard Wyss, Sebastian Schneeweiss, and Mark~J van~der Laan.
\newblock Propensity score prediction for electronic healthcare databases using super learner and high-dimensional propensity score methods.
\newblock \emph{Journal of Applied Statistics}, 46\penalty0 (12):\penalty0 2216--2236, 2019.

\bibitem[Khwaja et~al.(2015)Khwaja, Naeem, Anpalagan, Venetsanopoulos, and Venkatesh]{khwaja2015improved}
AS Khwaja, Muhammad Naeem, A Anpalagan, A Venetsanopoulos, and B Venkatesh.
\newblock Improved short-term load forecasting using bagged neural networks.
\newblock \emph{Electric Power Systems Research}, 125:\penalty0 109--115, 2015.

\bibitem[Kilimci and Akyoku{\c{s}}(2018)]{kilimci2018deep}
Zeynep~Hilal Kilimci and Selim Akyoku{\c{s}}.
\newblock Deep learning-and word embedding-based heterogeneous classifier ensembles for text classification.
\newblock \emph{Complexity}, 2018.

\bibitem[Kuznetsov et~al.(2014)Kuznetsov, Mohri, and Syed]{kuznetsov2014multi}
Vitaly Kuznetsov, Mehryar Mohri, and Umar Syed.
\newblock Multi-class deep boosting.
\newblock \emph{Advances in Neural Information Processing Systems}, 27, 2014.

\bibitem[Kwak and Park(2022)]{kwak2022unsupervised}
Geun-Ho Kwak and No-Wook Park.
\newblock Unsupervised domain adaptation with adversarial self-training for crop classification using remote sensing images.
\newblock \emph{Remote Sensing}, 14\penalty0 (18):\penalty0 4639, 2022.

\bibitem[Liu et~al.(2014)Liu, Han, Meng, and Tong]{liu2014facial}
Ping Liu, Shizhong Han, Zibo Meng, and Yan Tong.
\newblock Facial expression recognition via a boosted deep belief network.
\newblock In \emph{Proceedings of the IEEE conference on computer vision and pattern recognition}, pages 1805--1812, 2014.

\bibitem[Lu et~al.(2022)Lu, Chen, Olaniyi, and Huang]{lu2022generative}
Yuzhen Lu, Dong Chen, Ebenezer Olaniyi, and Yanbo Huang.
\newblock Generative adversarial networks (gans) for image augmentation in agriculture: A systematic review.
\newblock \emph{Computers and Electronics in Agriculture}, 200:\penalty0 107208, 2022.

\bibitem[Ma et~al.(2019)Ma, Deng, Qi, Jiang, Li, Wang, and Xing]{ma2019fully}
Xu Ma, Xiangwu Deng, Long Qi, Yu Jiang, Hongwei Li, Yuwei Wang, and Xupo Xing.
\newblock Fully convolutional network for rice seedling and weed image segmentation at the seedling stage in paddy fields.
\newblock \emph{PloS one}, 14\penalty0 (4):\penalty0 e0215676, 2019.

\bibitem[Ma and Zhang(2022)]{ma2022multi}
Yuchi Ma and Zhou Zhang.
\newblock Multi-source unsupervised domain adaptation on corn yield prediction.
\newblock In \emph{AI for Agriculture and Food Systems}, 2022.

\bibitem[Mulla and Khosla(2016)]{mulla2016historical}
David Mulla and Raj Khosla.
\newblock Historical evolution and recent advances in precision farming.
\newblock \emph{Soil-specific farming precision agriculture}, pages 1--35, 2016.

\bibitem[Ronneberger et~al.(2015)Ronneberger, Fischer, and Brox]{ronneberger2015u}
Olaf Ronneberger, Philipp Fischer, and Thomas Brox.
\newblock {U-Net}: Convolutional networks for biomedical image segmentation.
\newblock In \emph{International Conference on Medical image computing and computer-assisted intervention}, pages 234--241. Springer, 2015.

\bibitem[Sapkota et~al.(2022)Sapkota, Hu, and Bagavathiannan]{sapkota2022evaluating}
Bishwa~B Sapkota, Chengsong Hu, and Muthukumar~V Bagavathiannan.
\newblock Evaluating cross-applicability of weed detection models across different crops in similar production environments.
\newblock \emph{Frontiers in Plant Science}, 13:\penalty0 837726, 2022.

\bibitem[Shkanaev et~al.(2020)Shkanaev, Sholomov, and Nikolaev]{shkanaev2020unsupervised}
Aleksandr~Yu Shkanaev, Dmitry~L Sholomov, and Dmitry~P Nikolaev.
\newblock Unsupervised domain adaptation for dnn-based automated harvesting.
\newblock In \emph{Twelfth International Conference on Machine Vision (ICMV 2019)}, pages 243--249. SPIE, 2020.

\bibitem[Simonyan and Zisserman(2014)]{simonyan2014very}
Karen Simonyan and Andrew Zisserman.
\newblock Very deep convolutional networks for large-scale image recognition.
\newblock \emph{arXiv preprint arXiv:1409.1556}, 2014.

\bibitem[Su et~al.(2021)Su, Kong, Qiao, and Sukkarieh]{su2021data}
Daobilige Su, He Kong, Yongliang Qiao, and Salah Sukkarieh.
\newblock Data augmentation for deep learning based semantic segmentation and crop-weed classification in agricultural robotics.
\newblock \emph{Computers and Electronics in Agriculture}, 190:\penalty0 106418, 2021.

\bibitem[Tao et~al.(2006)Tao, Tang, Li, and Wu]{tao2006asymmetric}
Dacheng Tao, Xiaoou Tang, Xuelong Li, and Xindong Wu.
\newblock Asymmetric bagging and random subspace for support vector machines-based relevance feedback in image retrieval.
\newblock \emph{IEEE Transactions on Pattern Analysis and Machine Intelligence}, 28\penalty0 (7):\penalty0 1088--1099, 2006.

\bibitem[Tranheden et~al.(2021)Tranheden, Olsson, Pinto, and Svensson]{tranheden2021dacs}
Wilhelm Tranheden, Viktor Olsson, Juliano Pinto, and Lennart Svensson.
\newblock Dacs: Domain adaptation via cross-domain mixed sampling.
\newblock In \emph{Proceedings of the IEEE/CVF Winter Conference on Applications of Computer Vision}, pages 1379--1389, 2021.

\bibitem[Ullah et~al.(2021)Ullah, Asad, and Bais]{ullah2021end}
Hafiz~Sami Ullah, Muhammad~Hamza Asad, and Abdul Bais.
\newblock End to end segmentation of canola field images using dilated {U-Net}.
\newblock \emph{IEEE Access}, 9:\penalty0 59741--59753, 2021.

\bibitem[Walach and Wolf(2016)]{walach2016learning}
Elad Walach and Lior Wolf.
\newblock Learning to count with {CNN} boosting.
\newblock In \emph{Computer Vision--ECCV 2016: 14th European Conference, Amsterdam, The Netherlands, October 11-14, 2016, Proceedings, Part II 14}, pages 660--676. Springer, 2016.

\bibitem[Wang et~al.(2020{\natexlab{a}})Wang, Xu, Wei, and Cui]{wang2020semantic}
Aichen Wang, Yifei Xu, Xinhua Wei, and Bingbo Cui.
\newblock Semantic segmentation of crop and weed using an encoder-decoder network and image enhancement method under uncontrolled outdoor illumination.
\newblock \emph{Ieee Access}, 8:\penalty0 81724--81734, 2020{\natexlab{a}}.

\bibitem[Wang et~al.(2020{\natexlab{b}})Wang, Xue, and Zhang]{wang2020particle}
Bin Wang, Bing Xue, and Mengjie Zhang.
\newblock Particle swarm optimisation for evolving deep neural networks for image classification by evolving and stacking transferable blocks.
\newblock In \emph{2020 IEEE Congress on Evolutionary Computation (CEC)}, pages 1--8. IEEE, 2020{\natexlab{b}}.

\bibitem[Wang et~al.(2022)Wang, Cao, Zhang, Li, Xu, and Wu]{wang2022review}
Dashuai Wang, Wujing Cao, Fan Zhang, Zhuolin Li, Sheng Xu, and Xinyu Wu.
\newblock A review of deep learning in multiscale agricultural sensing.
\newblock \emph{Remote Sensing}, 14\penalty0 (3):\penalty0 559, 2022.

\bibitem[Welchowski and Schmid(2016)]{welchowski2016framework}
Thomas Welchowski and Matthias Schmid.
\newblock A framework for parameter estimation and model selection in kernel deep stacking networks.
\newblock \emph{Artificial Intelligence in Medicine}, 70:\penalty0 31--40, 2016.

\bibitem[Wolpert(1992)]{wolpert1992stacked}
David~H Wolpert.
\newblock Stacked generalization.
\newblock \emph{Neural Networks}, 5\penalty0 (2):\penalty0 241--259, 1992.

\bibitem[Wu et~al.(2020)Wu, Sahoo, and Hoi]{wu2020recent}
Xiongwei Wu, Doyen Sahoo, and Steven~CH Hoi.
\newblock Recent advances in deep learning for object detection.
\newblock \emph{Neurocomputing}, 396:\penalty0 39--64, 2020.

\bibitem[Wu et~al.(2018)Wu, Han, Lin, Uzunbas, Goldstein, Lim, and Davis]{wu2018dcan}
Zuxuan Wu, Xintong Han, Yen-Liang Lin, Mustafa~Gokhan Uzunbas, Tom Goldstein, Ser~Nam Lim, and Larry~S Davis.
\newblock Dcan: Dual channel-wise alignment networks for unsupervised scene adaptation.
\newblock In \emph{Proceedings of the European Conference on Computer Vision (ECCV)}, pages 518--534, 2018.

\bibitem[Yu et~al.(2022)Yu, Che, Yu, and Zhang]{yu2022development}
Helong Yu, Minghang Che, Han Yu, and Jian Zhang.
\newblock Development of weed detection method in soybean fields utilizing improved deeplabv3+ platform.
\newblock \emph{Agronomy}, 12\penalty0 (11):\penalty0 2889, 2022.

\bibitem[Zhang et~al.(2020)Zhang, Jiang, Shao, and Cui]{zhang2020snapshot}
Wentao Zhang, Jiawei Jiang, Yingxia Shao, and Bin Cui.
\newblock Snapshot boosting: a fast ensemble framework for deep neural networks.
\newblock \emph{Science China Information Sciences}, 63:\penalty0 1--12, 2020.

\bibitem[Zheng and Yang(2019)]{zheng2019unsupervised}
Zhedong Zheng and Yi Yang.
\newblock Unsupervised scene adaptation with memory regularization in vivo.
\newblock \emph{arXiv preprint arXiv:1912.11164}, 2019.

\bibitem[Zheng and Yang(2021)]{zheng2021rectifying}
Zhedong Zheng and Yi Yang.
\newblock Rectifying pseudo label learning via uncertainty estimation for domain adaptive semantic segmentation.
\newblock \emph{International Journal of Computer Vision}, 129\penalty0 (4):\penalty0 1106--1120, 2021.

\bibitem[Zou et~al.(2021)Zou, Chen, Wang, Zhang, and Zhang]{zou2021modified}
Kunlin Zou, Xin Chen, Yonglin Wang, Chunlong Zhang, and Fan Zhang.
\newblock A modified u-net with a specific data argumentation method for semantic segmentation of weed images in the field.
\newblock \emph{Computers and Electronics in Agriculture}, 187:\penalty0 106242, 2021.

\end{thebibliography}
